\documentclass{article}

\usepackage[numbers,square,sort&compress]{natbib}

\usepackage[preprint]{neurips_2026}

\usepackage[utf8]{inputenc} 
\usepackage[T1]{fontenc}    
\usepackage[hidelinks]{hyperref}  
\usepackage{url}            
\usepackage{booktabs}       
\usepackage{amsfonts}       
\usepackage{nicefrac}       
\usepackage{microtype}      
\usepackage{xcolor}         
\usepackage{graphicx}

\usepackage{amsmath}
\usepackage{amssymb}
\usepackage{mathtools}
\usepackage{amsthm}
\usepackage{makecell}
\usepackage{comment}
\usepackage[capitalize,noabbrev]{cleveref}

\usepackage{subcaption}
\usepackage{lipsum}
\usepackage{algorithm}
\usepackage{algpseudocode}
\usepackage{algorithmicx}
\usepackage{bm}
\usepackage{wrapfig}
\usepackage[super]{nth}

\usepackage[frozencache,cachedir=minted-cache]{minted}
\setminted{
    style=borland,
    linenos,
    numbersep=4pt,
    xleftmargin=1em,
}

\usepackage[acronym,toc,shortcuts]{glossaries}
\glsdisablehyper
\newacronym{ukb}{UKB}{UK Biobank}
\newacronym{sota}{sota}{state-of-the-art}
\newacronym{flops}{FLOPS}{Floating Point Operations}
\newacronym{ce}{CE}{Cross Entropy}
\newacronym{bce}{BCE}{Binary Cross Entropy}
\newacronym{ecg}{ECG}{electrocardiogram}
\newacronym{ehr}{EHR}{electronic health record}
\newacronym{mri}{MRI}{Magnetic Resonance Images}
\newacronym{ct}{CT}{computer-assisted tomography}
\newacronym{prs}{PRS}{polygenic risk score}
\newacronym{auroc}{AUROC}{area under the receiver operating characteristic}
\newacronym{vit}{ViT}{Vision Transformer}
\newacronym{mlp}{MLP}{Multi-Layer Perceptron}
\newacronym{cnn}{CNN}{Convolutional Neural Network}
\newacronym{vlm}{VLM}{Vision-Language Models}
\newacronym{sd}{SD}{standard deviation}
\newacronym{se}{SE}{standard error}
\newacronym{gpu}{GPU}{Graphics Processing Unit}
\newacronym{mip}{MIP}{multilinear inner product}
\newacronym{tc}{TC}{total correlation}
\newacronym{moe}{MoE}{mixture-of-experts model}
\newacronym{poe}{PoE}{product-of-experts model}
\newacronym{ood}{OOD}{Out-of-Distribution}
\newacronym{cam}{CAM}{class activation map}

\newcommand{\ie}{\textit{, i.e., }}
\newcommand{\eg}{\textit{, e.g., }}

\title{
    Hidden in the Multiplicative Interaction: Uncovering Fragility in Multimodal Contrastive Learning
}

\author{
  \textbf{Tillmann Rheude\textsuperscript{1,3$^\dagger$}}, 
  \textbf{Stefan Hegselmann\textsuperscript{1}}, 
  \textbf{Roland Eils\textsuperscript{1,2,3$^\dagger$}}, 
  \textbf{Benjamin Wild\textsuperscript{1$^\dagger$}}\\
  \textsuperscript{1}Berlin Institute of Health, Charité - Universitätsmedizin Berlin, 
  \textsuperscript{2}Intelligent Medicine Institute,\\Fudan University, 
  \textsuperscript{3}Department of Mathematics and Computer Science, Freie Universität Berlin\vspace{2mm} \\
  \textsuperscript{$^\dagger$}\texttt{\{tillmann.rheude, roland.eils, benjamin.wild\}@bih-charite.de}
}

\begin{document}

\maketitle

\begin{abstract}
    Contrastive learning has become a standard approach for unsupervised learning from paired data, as demonstrated by CLIP for image-text matching. However, many domains involve more than two modalities and require objectives that capture higher-order dependencies beyond pairwise alignment. Symile extends CLIP to this setting by replacing the dot product with the \ac{mip} over modality embeddings. 
    In this work, we show that there is a fragility which is \emph{hidden in the multiplicative interaction}: a single weakly informative, misaligned, or missing modality can propagate through the objective and distort cross-modal retrieval scores.
    We propose Gated Symile, a contrastive gating mechanism that adapts modality contributions on an attention-based, per-candidate basis. 
    The gate suppresses unreliable inputs by interpolating embeddings toward learnable neutral directions with an explicit \texttt{NULL} option when reliable cross-modal alignment is unlikely.
    Across a controlled synthetic benchmark that \emph{uncovers this fragility} and three real-world trimodal datasets, Gated Symile achieves higher top-1 retrieval accuracy than well-tuned \ac{sota} baselines.
    More broadly, our results highlight gating as a step toward robust multimodal contrastive learning beyond two modalities in the presence of noise, misalignment, or missing inputs.
    \footnote{The code repository is uploaded to \href{https://github.com/TillmannRheude/gated_symile}{GitHub}.}
\end{abstract}

\section{Introduction}
\label{sec:introduction}

Contrastive learning has become a standard tool for bimodal learning exemplified with image-text pairs \cite{Radford_Kim_Hallacy_Ramesh_Goh_Agarwal_Sastry_Askell_Mishkin_Clark_2021}.
However, many real-world problems require reasoning over more than two modalities.
For instance, medical prediction tasks often combine more modalities \cite{Acosta_Falcone_Rajpurkar_Topol_2022}, including imaging \cite{Tak_Garomsa_Zapaishchykova_Chaunzwa_ClimentPardo_Ye_Zielke_Ravipati_Pai_Vajapeyam_etal_2026}, laboratory measurements \cite{Saporta_Puli_Goldstein_Ranganath_2024}, proteomics \cite{Rheude_Eils_Wild_2025_CAMA}, metabolomics \cite{Buergel_Steinfeldt_Ruyoga_Pietzner_Bizzarri_Vojinovic_UpmeierZuBelzen_Loock_Kittner_Christmann_etal_2022}, \acp{ecg}, and \acp{ehr} \cite{Steinfeldt_Wild_Buergel_Pietzner_UpmeierZuBelzen_Vauvelle_Hegselmann_Denaxas_Hemingway_Langenberg_etal_2025}.
Hence, recent work extends bimodal contrastive learning with objectives that model higher-order interactions across modalities \cite{Saporta_Puli_Goldstein_Ranganath_2024,Cicchetti_Grassucci_Comminiello_2025,cicchetti2025gramian,Dufumier_Navarro_Tuia_Thiran_2025}. 
The objective for multimodal contrastive learning beyond bimodality could be\eg a triangle \cite{Cicchetti_Grassucci_Comminiello_2025} or a parallelotope \cite{cicchetti2025gramian} between modality embeddings to model all interactions instead of only pairwise interactions.
These can be generalized with Symile \cite{Saporta_Puli_Goldstein_Ranganath_2024} which extends CLIP's dot product to multimodality with the \ac{mip} as the contrastive objective.

However, real-world datasets often contain modalities that may be complementary, conflicting, weak, or missing.
For example, a patient with early-stage pneumonia may show signs in chest X-rays and lab tests, while the clinical notes do not yet document respiratory symptoms, leading to conflicting cross-modal signals for the same patient.
We find that existing multimodal contrastive learning methods treat all modalities symmetrically and cannot explicitly model reliability differences.
This results in a distortion of the cross-modal objective for \ac{sota} baselines.
Therefore, a central challenge in multimodal contrastive learning is to exploit higher-order interactions when modalities are informative, while remaining robust to weak, noisy, or missing inputs.

In this paper, we highlight an under-explored limitation of multimodal contrastive learning: weak or misaligned modalities can propagate through the objective and distort the learned signal. 
While one might expect modality encoders to circumvent this, we show that this is not guaranteed.
For CLIP's multimodal generalization\ie Symile, this effect is \emph{hidden in the multiplicative interaction}.
Because the \ac{mip} combines modalities multiplicatively, errors from a single modality are amplified by the others.
As a result, aggregate performance can remain relatively strong even though individual samples are affected by unreliable modalities, masking architectural fragility and leading to silent failures when modalities are misaligned or uninformative.
\begin{figure}[t]
    \centering
    \includegraphics[width=1.0\linewidth]{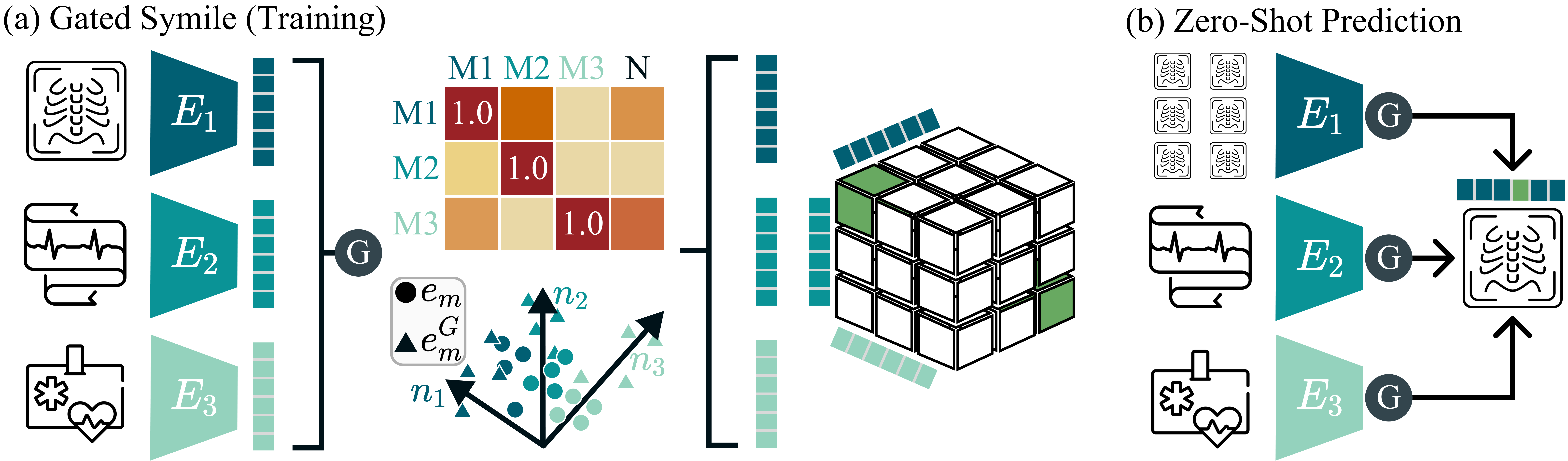}
    \caption{
        Illustrative overview of Gated Symile exemplified with the trimodal Symile-MIMIC \cite{Saporta_Puli_Goldstein_Ranganath_2024} dataset. 
        \textbf{(a)} During training, modality-specific encoders $E_m$ create embeddings $e_m$ and our proposed gate $G$ produces target‑conditioned weights over the available modalities and a \texttt{NULL} option $N$ (heatmap). The gate forms gated embeddings by weighting and interpolating each modality embedding with a modality‑specific neutral direction $n_m$ (coordinate system), enabling the model to suppress unreliable modalities while preserving useful signal. The gated embeddings are then combined via Symile's \ac{mip} (cube with positives on the diagonal). 
        \textbf{(b)} At inference, the gating is applied to compute candidate scores for zero‑shot prediction.
    }
    \label{fig:overview}
\end{figure}
Motivated by this observation, we introduce a gating mechanism on top of Symile to adaptively modulate modality contributions. 
The resulting method, Gated Symile (\Cref{fig:overview}), computes candidate-conditioned gate weights to attenuate unreliable modalities.
The gate interpolates misaligned embeddings toward neutral directions, with an explicit \texttt{NULL} option when the target embedding indicates that reliable cross-modal alignment is unlikely.
We find that Gated Symile strongly improves the robustness for misaligned modalities while preserving Symile's original higher-order interactions.

Taken together, our approach improves multimodal contrastive learning with modality-specific reliability and higher-order interactions.
Our key contributions are:
\begin{itemize}
    \item We identify and theoretically derive that \textbf{Symile's \ac{mip} implicitly treats all modalities symmetrically}, which can hide fragility\eg under modality misalignment.
    
    \item We propose Gated Symile, 
    \textbf{an attention-based per-candidate gating mechanism} that downweights unreliable modalities by interpolating embeddings toward learnable neutral directions and incorporating a \texttt{NULL} option for uninformative evidence.
    
    \item We propose \textbf{a new synthetic benchmark} with controlled modality misalignment, enabling systematic analysis of robustness to unreliable inputs.

    \item We demonstrate on synthetic and three real-world datasets that \textbf{Gated Symile consistently improves top-1 retrieval accuracy} over well-tuned \ac{sota} baselines.

    \item Beyond retrieval performance, we \textbf{analyze gate weights, embedding geometries, scaling behavior, model efficiency, and ablate gate features} to better understand the impact of modality misalignment and the performance gains enabled by gating.
\end{itemize}

\section{Related Work}
\label{sec:related_work}
\paragraph{Contrastive Learning}
Contrastive learning has become a dominant paradigm for representation learning.
Its optimizing objective pulls together embeddings of related views while pushing unrelated ones apart.
In the unimodal setting, this is typically done with instance discrimination under data augmentations\eg with the InfoNCE objective \cite{Oord_Li_Vinyals_2019}, SimCLR \cite{Chen_Kornblith_Norouzi_Hinton_2020}, and MoCo \cite{He_Fan_Wu_Xie_Girshick_2020}. 
Bimodal contrastive learning extends this to cross-modal representations, in which pairs are positives across modalities exemplified by image-text pairs in CLIP \cite{Radford_Kim_Hallacy_Ramesh_Goh_Agarwal_Sastry_Askell_Mishkin_Clark_2021} and follow-ups such as SigLIP \cite{Zhai_Mustafa_Kolesnikov_Beyer_2023}. 
Moving beyond two modalities, multimodal methods can benefit from objectives enforcing agreement across all modalities\eg by matching relational structure across modalities in addition to instance-level pairing.
This is exemplified with methods like AudioClip \cite{Guzhov_Raue_Hees_Dengel_2022}, ImageBind \cite{Girdhar_ElNouby_Liu_Singh_Alwala_Joulin_Misra_2023}, Gram \cite{cicchetti2025gramian}, Triangle \cite{Cicchetti_Grassucci_Comminiello_2025}, CoMM \cite{Dufumier_Navarro_Tuia_Thiran_2025}, and Symile \cite{Saporta_Puli_Goldstein_Ranganath_2024}. 
However, prior methods do not explicitly model robustness to varying modality interactions. In such cases, naive alignment objectives may be suboptimal when one modality is weakly informative or noisy.

\paragraph{Gating Mechanism}
Gating modulates information flow by selecting, reweighting, or routing representations.
Early and widely used examples include gates in recurrent networks such as LSTMs \cite{Hochreiter_Schmidhuber_1997} and GRUs \cite{Cho_Merrienboer_Bahdanau_Bengio_2014}, which regulate how much past state is retained and how new evidence is incorporated.
Beyond sequence models, gating is often used for conditional feature modulation.
For example, FiLM \cite{Perez_Strub_DeVries_Dumoulin_Courville_2018} applies conditioning-dependent scaling and shifting to intermediate activations.
In Transformers \cite{Vaswani_Shazeer_Parmar_Uszkoreit_Jones_Gomez_Kaiser_Polosukhin_2017}, attention weights similarly implement soft selection by controlling how strongly tokens contribute to representations.
Channel- and spatial-wise gating has also been used, most prominently in Squeeze-and-Excitation blocks \cite{Hu_Shen_Sun_2018}, which learn per-channel importance weights to emphasize informative feature maps. 
Further, routing-based gates enable conditional computation by selecting subsets of expert modules, as in \acp{moe} \cite{Jordan_Jacobs_1994,Jacobs_Jordan_Nowlan_Hinton_1991} and \acp{poe} \cite{Hinton_2002}.
Most closely related to our setting is contrastive gating such as CDG \cite{Meng_Yang_Shin_Fan_Seo_2022}, CR-\ac{moe} \cite{Jiang_Zheng_Cheng_Awadallah_Wang_2024}, and MCMR \cite{Lu_Li_Huang_Meng_Zeng_Shen_2026} to suppress less informative inputs \cite{Wang_Liu_Li_Sheng_Yan_Wang_Shao_2019,Zohra_Zhao_Itani_Ghanem_2026,Wan_Wang_Stengel_Eskin_Cho_Bansal_2025,Gorti_Vouitsis_Ma_Golestan_Volkovs_Garg_Yu_2022}. 
We apply gating (i) within a contrastive objective, (ii) beyond bimodal settings, and (iii) explicitly to address modality misalignment.
This combination has not been studied in prior work.

\paragraph{Selective Prediction and Explicit Abstention}
Early work formalized selective classification\ie prediction with a reject option, as a principled risk-coverage trade-off \cite{Chow_1970,El-Yaniv_Wiener_2010}, and later instantiated it for deep models such as SelectiveNet \cite{Geifman_El-Yaniv_2019}.
Further related work include open set recognition\ie prediction with unknowns at test time \cite{Scheirer_deRezendeRocha_Sapkota_Boult_2013}, exemplified by architectures such as OpenMax \cite{Bendale_Boult_2016}.
This intersects with \ac{ood} detection, where abstention is often implemented through an explicit \texttt{NULL} pathway or confidence-based rejection \cite{Hendrycks_Gimpel_2017,Liang_Li_Srikant_2018}. 
In contrast, in our work, selection and rejection is represented more locally\eg as a probability mass and without supervision. 
This connects to the broader idea of learned neutral placeholders with special tokens such as \texttt{CLS}, \texttt{MASK}, and \texttt{REG} in BERT \cite{Devlin_Chang_Lee_Toutanova_2019} and \acp{vit} \cite{Dosovitskiy_Beyer_Kolesnikov_Weissenborn_Zhai_Unterthiner_Dehghani_Minderer_Heigold_Gelly_etal._2021,Darcet_Oquab_Mairal_Bojanowski_2024}, as well as learned prototypes such as discrete latents in VQ-VAEs \cite{vandenOord_Vinyals_kavukcuoglu_2017} and class prototypes in prototypical networks \cite{Snell_Swersky_Zemel_2017}.


\section{Method}
\label{sec:method}

\subsection{%
  \texorpdfstring
    {Sensitivity of the \ac{mip} to Modality Misalignment}
    {Sensitivity of the MIP to Modality Misalignment}
}
Symile maximizes a multi-sample contrastive lower bound on \ac{tc}, a measure of higher-order dependence among modalities \cite{Watanabe_1960,Saporta_Puli_Goldstein_Ranganath_2024}. For $M$ modalities,
\begin{equation}
\mathrm{TC}(x^{(1)},\dots,x^{(M)})
= D_{\mathrm{KL}}\!\left(p(x^{(1)},\dots,x^{(M)}) \,\middle\|\, \prod_{m=1}^M p(x^{(m)})\right),
\end{equation}
which is zero under mutual independence. Symile uses an InfoNCE-style objective to distinguish a positive tuple from negatives formed by sampling from the product of marginals, yielding a tractable lower bound on \ac{tc} with a learned critic $g$. To instantiate $g$, Symile replaces CLIP’s dot product with the \acrfull{mip}. Let $e_m = E_m(x^{(m)}) \in \mathbb{R}^{D}$ denote the embedding of modality $m$ with encoders $E_m$ and the shared embedding dimension $D$, then the \ac{mip} is
\begin{equation}
\langle e_1,\dots,e_M \rangle
= \sum_{j=1}^{D} \prod_{m=1}^M e_{m,j},
\end{equation}
and Symile scores tuples via $g(x^{(1)},\dots,x^{(M)})=\langle e_1,\dots,e_M\rangle/\tau_{\text{MIP}}$ with temperature $\tau_{\text{MIP}}$ \cite{Saporta_Puli_Goldstein_Ranganath_2024}.
For example, for a retrieval task with target modality $t$, misalignment in a single non-target modality can strongly distort scores because the \ac{mip} multiplies contributions across modalities:
\begin{equation}
g(x^{(1)},\dots,x^{(M)})
=
\frac{1}{\tau_{\text{MIP}}}\sum_{j=1}^{D}
\Bigg(e_{t,j}\prod_{\substack{m=1\\ m\neq t}}^{M} e_{m,j}\Bigg).
\end{equation}
If a non-target modality $c\neq t$ is perturbed\ie its embedding changes as $\hat e_c = e_c + \delta$, then
\begin{equation}
g_{\mathrm{corr}} - g_{\mathrm{clean}}
=
\frac{1}{\tau_{\text{MIP}}}\sum_{j=1}^{D}
e_{t,j}\Bigg(\prod_{\substack{m=1\\ m\neq t,c}}^{M} e_{m,j}\Bigg)\delta_j.
\label{eq:mip_corruption_delta}
\end{equation}
Thus the score error is linear in $\delta$ but scaled by the product of the remaining modalities. This perturbation does not assume any specific source and can reflect\eg misalignment to the ideal cross-modal tuple.
Rewriting this as an inner product and applying Cauchy-Schwarz yields
\begin{equation}
|g_{\mathrm{corr}} - g_{\mathrm{clean}}|
\le
\frac{1}{\tau_{\text{MIP}}}\,
\|\delta\|_2 \,
\Big\|\, e_t \odot \!\!\!\prod_{\substack{m=1\\ m\neq t,c}}^{M}\!\!\! e_m \Big\|_2,
\label{eq:mip_bound}
\end{equation}
highlighting how multiplicative interactions can cause perturbations from a single unreliable modality to scale with the remaining embeddings during both training and inference (Details in \Cref{app:cauchy_schwarz_derivation}).

\begin{wrapfigure}[16]{r}{0.25\textwidth}
        \vspace{-.4cm}
        \centering
        \includegraphics[width=0.8\linewidth]{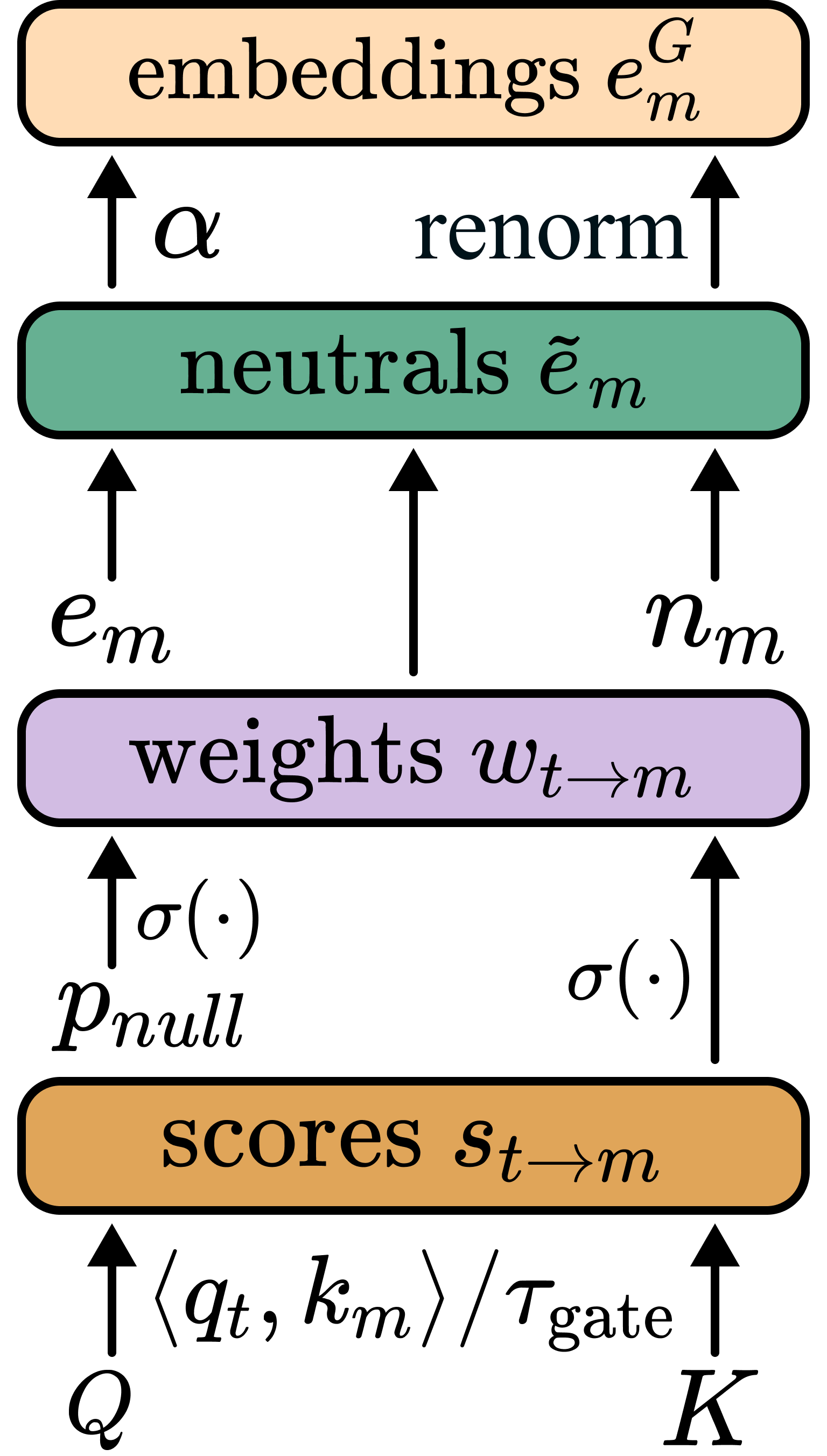}
        \caption{Sigmoid gate with \texttt{NULL} option and neutral directions.}
        \label{fig:illustrative_gate}
\end{wrapfigure}

\subsection{Gate Mechanism}
\label{sec:gate_mechanism}
To overcome the sensitivity of MIP, we introduce a gate that modulates the contribution of each modality in Symile’s \ac{mip}. 
For a retrieval direction, the gate outputs gated embeddings $e^G_1,\dots,e^G_M$ by using gate weights $\{w_{t\to m}\}_{m=1}^M$ that control how strongly each modality should influence the \ac{mip} score. 
Intuitively, the gate aims to suppress non-target modalities whose current sample provides unreliable evidence for the retrieval target\eg because the modality is misaligned, weakly informative, or missing.
The gate is illustrated in  \Cref{fig:illustrative_gate} and formalized in \Cref{alg:attention_gate}.
Embeddings $e_m$, projected queries/keys, neutral prototypes $n_m$, and gated embeddings $e_m^G$ are $\ell_2$-normalized.

\paragraph{Attention-Based, Candidate-Dependent Gating}
The proposed gate is candidate-dependent\ie the weights are computed conditioned on the target embedding $e_t$ and the candidate's non-target embeddings $\{e_m\}_{m\neq t}$.
Concretely, we form a query vector from the target modality, $q_t=Q_t(e_t)$, and key vectors for each non-target modality, $k_m = K_m(e_m)$ for $m\neq t$.
Due to $\ell_2$-normalization, the relevance score is a scaled cosine similarity where $\tau_{\text{gate}}>0$ controls the sharpness of the gating decisions\footnote{We use the same temperature $\tau_{\text{gate}}$ for the modality relevance scores and the \texttt{NULL} gating logit, so that $\tau_{\text{gate}}$ jointly controls the sharpness of both gating weights and the \texttt{NULL} decision.},
\begin{equation}
s_{t\to m} \;=\; \langle q_t, k_m\rangle / \tau_{\text{gate}},
\end{equation}
and is mapped to a gate weight with an activation function $\sigma$\eg a sigmoid or softmax function,
\begin{equation}
w_{t\to m} \;=\; \sigma(s_{t\to m}) \in (0,1).
\end{equation}
We set $w_{t\to t}=1$ so that the target modality is never suppressed. To disentangle the effect of candidate dependence from the act of reweighting itself, we also consider an ablation with a lightweight baseline that replaces attention scores with a learned static matrix of gating logits (per target-modality pair).

\paragraph{Neutral Directions}
We introduce a per-modality neutral prototype $n_m\in\mathbb{R}^D$ to make downweighting explicit in representation space. For each modality, we interpolate between the current embedding and its neutral direction:
\begin{equation}
\tilde e_m \;=\; w_{t\to m}\, e_m \;+\; (1-w_{t\to m})\, n_m.
\label{eq:neutral_direction_interpolation}
\end{equation}
Thus, a small $w_{t\to m}$ pushes a modality $m$ toward a learned neutral embedding, making its contribution to the \ac{mip} closer to a non-informative baseline rather than injecting noise.

\paragraph{Gate Strength and Renormalization}
We include a strength parameter $\alpha\in[0,1]$ that blends between the identity and the fully gated embedding analogous to a residual connection \cite{He_Zhang_Ren_Sun_2016}:
\begin{equation}
e_m^G \;=\; \!(1-\alpha)\,e_m + \alpha\,\tilde e_m.
\label{eq:gated_embedding}
\end{equation}
We $\ell_2$-normalize $e_m^G$ after gating to keep magnitudes comparable across settings and prevent the gate from trivially changing the score via norm scaling.
Hence, the gate primarily affects the direction of each modality embedding.

\begin{wrapfigure}[27]{R}{0.5\textwidth}
\begin{minipage}{0.5\textwidth}
\vspace{-0.9cm}
\begin{algorithm}[H]

\caption{Attention-based gate with sigmoid, \texttt{NULL} option, and neutral directions.} 
\label{alg:attention_gate}

\begin{algorithmic}[1] 
\Require $e_1,\dots,e_M \in \mathbb{R}^D$, target index $t$ 
\Require $Q_t:\mathbb{R}^D\!\to\!\mathbb{R}^{d_k}$, $K_m:\mathbb{R}^D\!\to\!\mathbb{R}^{d_k}$ 
\Require $h_t:\mathbb{R}^D\!\to\!\mathbb{R}$ and $u_t \in \mathbb{R}$ 
\Require $n_1,\dots,n_M \in \mathbb{R}^D$ 
\Require $\tau_\text{gate}>0$, $\alpha\in[0,1]$
\State $q_t \gets \mathrm{norm}(Q_t(e_t))$

\For{$m \in \{1,\dots,M\}\setminus\{t\}$} 
\State $k_m \gets \mathrm{norm}(K_m(e_m))$

\State $s_{t\to m} \gets \langle q_t, k_m\rangle / \tau_\text{gate}$
\State $w_{t\to m} \gets \sigma(s_{t\to m})$ \Comment{sigmoid weight} 
\EndFor 
\State $z_t \gets (h_t(e_t) + u_t)/\tau_\text{gate}$ 
\Comment{\texttt{NULL} logit} 
\State $p_{\mathrm{null}} \gets \sigma(z_t)$ 
\For{$m \neq t$} 
\State $w_{t\to m} \gets (1-p_{\mathrm{null}})\,w_{t\to m}$ 
\EndFor 
\State $w_{t\to t} \gets 1$ 

\For{$m=1$ to $M$} 
\State $\tilde{e}_m \gets w_{t\to m}\,e_m $
\Statex \hspace{\algorithmicindent} $\qquad + (1-w_{t\to m})n_m$ 
\Comment{neutral}

\State $e^G_m \gets (1-\alpha)\,e_m + \alpha\,\tilde{e}_m$ 
\Comment{gate strength} 

\State $e^G_m \gets \mathrm{norm}(e^G_m)$ \Comment{renorm} 
\EndFor 

\State \Return gated embeddings $e^G_1,\dots,e^G_M$ 

\end{algorithmic} 
\end{algorithm}

\end{minipage}
\end{wrapfigure}

\paragraph{Null Option}
Since our main gate uses independent sigmoid activations (and softmax as an ablation), multiple non-target modalities can be downweighted simultaneously.
We additionally include a \texttt{NULL} option in the gating mechanism (not an additional embedding in the \ac{mip}) to allow the model to down-weight cross-modal evidence when the target embedding indicates that reliable cross-modal alignment is unlikely.
Specifically, we compute a \texttt{NULL} logit with a projection head $h_t$ and a bias $u_t$ as 
\begin{equation}
z_t \;=\; \big(h_t(e_t)+u_t\big)/\tau_{\text{gate}},
\end{equation}
and set $p_{\mathrm{null}}=\sigma(z_t)$ for the sigmoid case, which multiplicatively shrinks all non-target weights by $(1-p_{\mathrm{null}})$.
Under the softmax gate, \texttt{NULL} is implemented as an additional logit category appended to the softmax; under the sigmoid gate, it is implemented as an independent probability shared across non-target modalities. 
In both cases, \texttt{NULL} affects the \ac{mip} only indirectly by suppressing non-target contributions, thereby pushing the corresponding gated embeddings toward their neutral directions.

\paragraph{Robustness of the Gated \ac{mip} to Modality Misalignment}
Analogous to the ungated case, if a non-target modality is perturbed, the pre-normalized gated embedding gets edited by the gate:
\begin{equation}
\hat e_c^G=e_c^G+\beta_{t\to c}\delta \hspace{1cm} \text{with } \hspace{1cm} \beta_{t\to c}=1-\alpha+\alpha(1-p_{\mathrm{null}})w_{t\to c}.
\label{eq:embedding_edited_by_gate}
\end{equation}
Therefore, the gated score satisfies
\begin{equation}
|g^G_{\mathrm{corr}} - g^G_{\mathrm{clean}}|
\le
\frac{\beta_{t\to c}}{\tau_{\text{MIP}}}\,
\|\delta\|_2 \,
\Big\|\, e_t^G \odot \!\!\!\prod_{\substack{m=1\\ m\neq t,c}}^{M}\!\!\! e_m^G \Big\|_2.
\end{equation}
Thus, when the gate suppresses an unreliable modality, the embedding entering the \ac{mip} is contracted by $\beta_{t\to c}$, reducing the multiplicative score distortion in \Cref{eq:mip_corruption_delta} (derivation in \Cref{app:gated_mip_sensitivity}).

\section{Experiments}
\label{sec:experiments}

\subsection{Datasets}
We focus on trimodal retrieval settings ($M=3$) analogous to prior work on multimodal contrastive learning \cite{Saporta_Puli_Goldstein_Ranganath_2024,Cicchetti_Grassucci_Comminiello_2025,cicchetti2025gramian,Dufumier_Navarro_Tuia_Thiran_2025}. 
We select datasets to probe complementary aspects (\Cref{tab:dataset_summary}).
Synthetic-\texttt{XNOR} isolates modality alignment in a controlled setting.
Symile-MIMIC \cite{Saporta_Puli_Goldstein_Ranganath_2024} provides a well-established, non-saturated benchmark.
The \ac{ukb} \cite{Sudlow_Gallacher_Allen_Beral_Burton_Danesh_Downey_Elliott_Green_Landray_etal_2015} enables evaluation at larger scale, while \ac{ukb}-Union further stresses the method under large-scale and missing-modality conditions.

\paragraph{Synthetic-\texttt{XNOR}}
We introduce a synthetic benchmark to study retrieval under controlled modality misalignment with a known ground-truth interaction.
The aim is to evaluate retrieval when one of the two non-target modalities is partly misleading.
Although one non-target modality remains informative, Symile’s \ac{mip} entangles both non-target signals, so a misaligned modality can dominate the interaction and prevent learning from the clean evidence.
We sample binary vectors $u,v\in\{0,1\}^{K}$ ($K=16$) with i.i.d.\ $\mathrm{Bernoulli}(0.5)$ bits and define the interaction $uv := \texttt{XNOR}(u,v)$.
The target modality encodes $A=[u,v,uv]$, while the non-target modalities encode complementary signals $B=[u,1,u]$ and $C=[1,v,v]$, so that for clean samples $B\odot C=[u,v,uv]$ matches the signal coordinates of $A$.
Each bit is embedded as $\{-s,+s\}$ on signal coordinates ($s=1$) and remaining dimensions contain Gaussian distractors ($\sigma=3$), producing a low signal-to-noise setting.
With probability $p$, we replace one modality in $\{B,C\}$ with the modality from another sample to simulate in-distribution misalignment.
The swapping preserves marginal statistics but breaks cross-modal alignment, preventing trivial noise detection which is an easy task for the encoders to learn.

\begin{table*}[t]
    \centering
    \caption{
        Overview of benchmark datasets for our evaluation. For the Synthetic-\texttt{XNOR} dataset, $u, v \in \{0,1\}^{K}$ with $K=16$, and $uv = \texttt{XNOR}(u,v)$ applied element-wise.
    }
    \label{tab:dataset_summary}
    \small
    \begin{tabular}{@{}l c c c@{}}
    \toprule
    \textbf{Dataset} & 
    \textbf{\# Samples} & 
    \textbf{Modalities} & 
    \textbf{Retrieval} \\
    \midrule

    Synthetic-\texttt{XNOR} 
        & $30,000$ & $A=[u,v,uv]$, $B=[u,1,u]$, $C=[1,v,v]$ & $A$ \\
        
    Symile-MIMIC \citep{Saporta_Puli_Goldstein_Ranganath_2024}
        & $10,345$ & Chest X-ray, Laboratory, \acs{ecg} & Chest X-ray  \\
        
    \acs{ukb} \citep{Sudlow_Gallacher_Allen_Beral_Burton_Danesh_Downey_Elliott_Green_Landray_etal_2015}
        & $37,888$
        & Proteomics, Metabolomics, \acs{ehr}
        & Proteomics \\

    \acs{ukb}-Union \citep{Sudlow_Gallacher_Allen_Beral_Burton_Danesh_Downey_Elliott_Green_Landray_etal_2015}
        & $486,400$
        & Proteomics, Metabolomics, \acs{ehr}
        & Proteomics \\
        
    \bottomrule
\end{tabular}
\end{table*}

\paragraph{Symile-MIMIC}
    The Symile-MIMIC dataset \cite{Saporta_Puli_Goldstein_Ranganath_2024} comprises $10,345$ samples collected from patients in an intensive care unit.
    The dataset contains three modalities: laboratory tests, chest X-ray images, and \acp{ecg}.
    The retrieval task is set up for the most expensive modality\ie the chest X-rays (\Cref{fig:overview}b), so the evaluation can be interpreted as zero-shot prediction or prioritization of an expensive target modality from cheaper complementary evidence. 
    For the whole setup including the actual implementation of the retrieval task and the encoders (\ac{mlp} for laboratory tests, ResNets \cite{He_Zhang_Ren_Sun_2016} for the vision and \acp{ecg} modalities), we follow \citet{Saporta_Puli_Goldstein_Ranganath_2024}.

\paragraph{\acf{ukb}}  
    The \ac{ukb} \cite{Sudlow_Gallacher_Allen_Beral_Burton_Danesh_Downey_Elliott_Green_Landray_etal_2015} is a large prospective biomedical cohort including a diverse range of modalities and possible tasks.
    We focus on proteomics, metabolomics and \acp{ehr} for the modalities and on a retrieval task analogous to the other datasets.
    We choose proteomics to be retrieved, since this represents one of the most expensive modalities for acquisition \citep{Rheude_Eils_Wild_2025_CAMA}.
    We use both the intersection of modalities\ie no missing modalities and $37,888$ samples, and the union of modalities\ie missing modalities and $486,400$ samples.
    Missing modalities are implemented analogous to \citet{Saporta_Puli_Goldstein_Ranganath_2024} by appending a binary mask indicating missingness to the input modality.
    We use normalized, raw modality inputs except for the \ac{ehr} modality.
    Here, we use QWEN \citep{Zhang_Li_Long_Zhang_Lin_Yang_Xie_Yang_Liu_Lin_etal_2025} embeddings \citep{Hegselmann_Arnim_Rheude_Kronenberg_Sontag_Hindricks_Eils_Wild_2025}.
    Modalities are encoded with \acp{mlp}.

\subsection{Experimental Setup}
We follow best practices for multimodal learning \cite{Rheude_Eils_Wild_2025} with consistent optimizer choices, coherent initializations, and hyperparameter tuning (\Cref{app:hyperparameter_tuning}).
For non-synthetic datasets (\ac{ukb} and Symile-MIMIC), we use $5$-fold cross-validation and report mean $\pm$ \ac{se} over three random seeds per fold.
For Synthetic-\texttt{XNOR}, we use a fixed train, validation, and test split.

\paragraph{Optimization}
We use ScheduleFree-AdamW \cite{Defazio_Yang_Khaled_Mishchenko_Mehta_Cutkosky_2024} and apply gradient clipping to stabilize training.
We use a learned logit scale $s=\exp(\gamma)$ to control the softmax temperature, and for Symile-style objectives we additionally apply a fixed $(d,M)$-dependent normalization to the \ac{mip} before the learned scaling to stabilize training across embedding dimensions and numbers of modalities.
We optimize gate parameters jointly with the encoders but use a separate learning rate multiplier for the gate module parameters.
Details are in \Cref{app:additionals}.

\paragraph{Sampling} Symile supports two negative-sampling regimes: $n$ (shuffled negatives) and $n^2$ (all pairings) \cite{Saporta_Puli_Goldstein_Ranganath_2024}.
In both cases, negatives are defined over combinations of the non-target modalities: in $n$-sampling, these are mismatched batch tuples, whereas in $n^2$-sampling all pairwise combinations are considered.
However, we introduce candidate-dependent scoring and gating to Symile.
To make this tractable, we use a $pair$ formulation inspired by vision-language pretraining \cite{Li_Selvaraju_Gotmare_Joty_Xiong_Hoi_2021,Bao_Wang_Dong_Liu_Mohammed_Aggarwal_Som_Piao_Wei_2022,Li_Li_Savarese_Hoi_2023,Li_Li_Xiong_Hoi_2022,Yu_Wang_Vasudevan_Yeung_Seyedhosseini_Wu_2022}.
Here, the target modality alone is varied while the remaining modalities are held fixed.
 For each query, this yields a candidate set consisting of the true positive and $K$ uniformly sampled negatives from the target modality ($K=128$).
Therefore, the gate is recomputed only for sampled target candidates rather than for all candidate combinations.
When applicable, methods are trained and evaluated with the same $pair$-based approximation to ensure a fair comparison.
We report $n$-sampling results separately in the ablation study.
Based on preliminary scaling experiments, we fix the batch size to $128$ (Synthetic-\texttt{XNOR}), $280$ (Symile-MIMIC, analogous to \citet{Saporta_Puli_Goldstein_Ranganath_2024}), and $512$ (\ac{ukb}).

\begin{table}[t]
    \centering
    \caption{
        Comparison of Gated Symile with well-tuned \ac{sota} baselines on synthetic ($p=1.0$) and real-world datasets. Values represent top-1 accuracy of the retrieval task (mean $\pm$ \ac{se}).
    }
    \label{tab:results}
    \small
    \begin{tabular}{@{}l c c c c@{}}
    
    \toprule
    
    \textbf{Method} & 
    \textbf{Synthetic-}\texttt{XNOR} {$\uparrow$} &
    \textbf{Symile-Mimic $\uparrow$} & 
    \textbf{\ac{ukb} $\uparrow$} &
    \textbf{\ac{ukb}-Union $\uparrow$} \\
    
    \midrule

    CLIP \cite{Radford_Kim_Hallacy_Ramesh_Goh_Agarwal_Sastry_Askell_Mishkin_Clark_2021}
        & $0.2434$ 
        & $0.4103 \pm 0.016$ 
        & $0.4089 \pm 0.015$
        & $0.0516 \pm 0.007$ \\

    TRIANGLE \cite{Cicchetti_Grassucci_Comminiello_2025}
        & $0.6093$ 
        & $0.0948 \pm 0.003$ 
        & $0.5651 \pm 0.012$ 
        & $0.3597 \pm 0.020$ \\

    GRAM \cite{cicchetti2025gramian}
        & $0.4864$ 
        & $0.2516 \pm 0.015$ 
        & $0.1848 \pm 0.007$ 
        & $0.2008 \pm 0.014$ \\

        
    Symile \cite{Saporta_Puli_Goldstein_Ranganath_2024}
        & $0.3310$ 
        & $0.4556 \pm 0.006$  
        & $0.6570 \pm 0.012$ 
        & $0.5278 \pm 0.009$ \\

    \midrule
        
    Gated Symile
        & $\mathbf{0.8733}$ 
        & $\mathbf{0.4670 \pm 0.005}$  
        & $\mathbf{0.6819 \pm 0.010}$
        & $\mathbf{0.6000 \pm 0.007}$  \\

    \bottomrule
\end{tabular}
\end{table}
\begin{figure}[t]
  \centering
  
  \begin{subfigure}{0.48\linewidth}
    \centering
    \includegraphics[width=\linewidth]{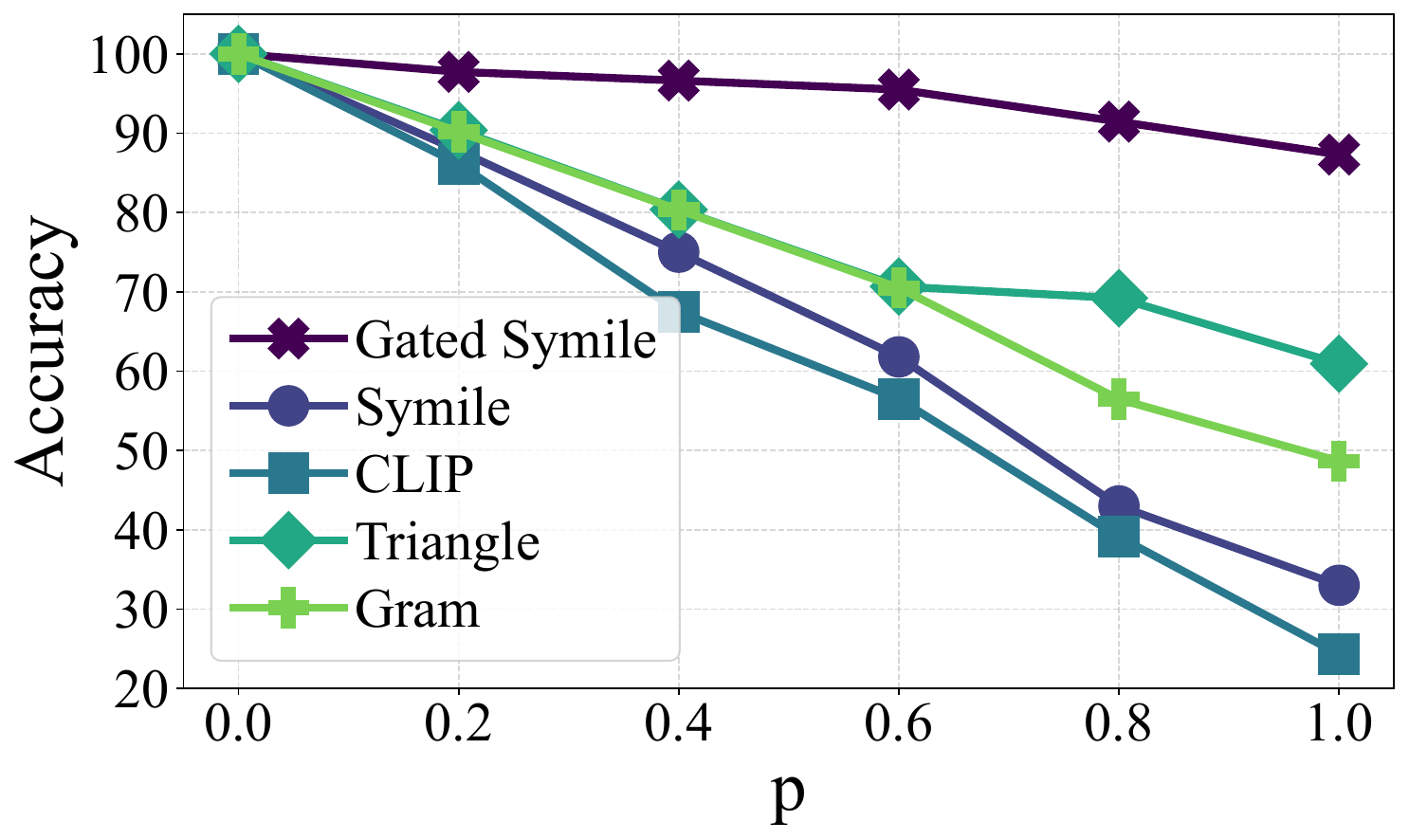}
  \end{subfigure}\hfill
  \begin{subfigure}{0.498\linewidth}
    \centering
    \includegraphics[width=\linewidth]{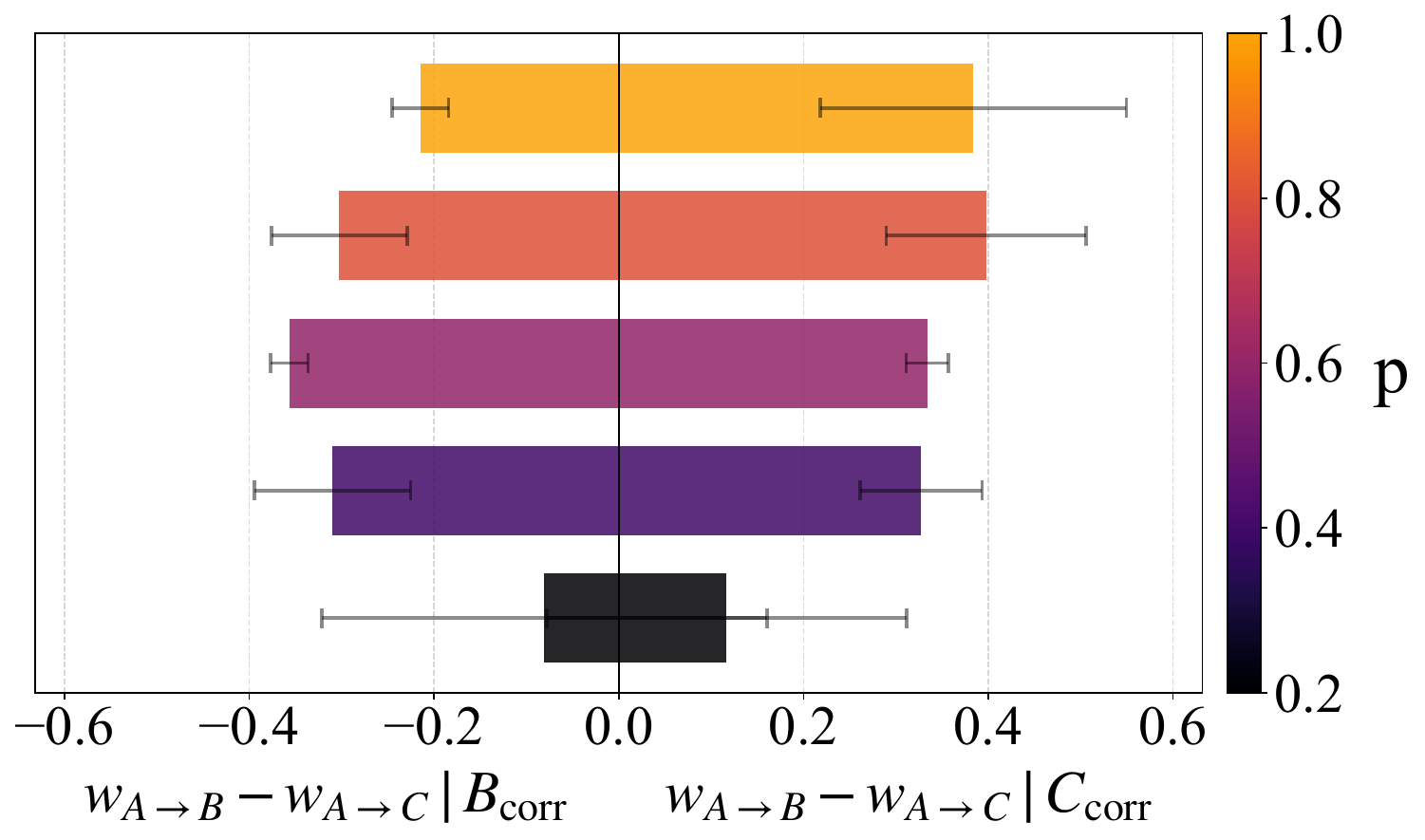}
  \end{subfigure}
  
  \caption{
        Analyses of well-tuned models on the Synthetic-\texttt{XNOR} dataset with probability $p$ of one non-target modality being misaligned.
        \textbf{(Left)} Decreasing retrieval accuracy under increasing misalignment. Symile only slightly outperforms CLIP along different values for $p$. The proposed Gated Symile, preserves the accuracy of Symile and prevents a collapse of the \ac{mip}, demonstrating that gating improves robustness to misaligned modalities. 
        \textbf{(Right)} Gate selects the reliable modality under misalignment. With two separate bar charts (toward left and right), we report the mean gate weight difference $w_{A\rightarrow B}-w_{A\rightarrow C}$. When $B$ is misaligned (left), the difference becomes negative\ie the gate assigns a smaller weight to $B$ than to $C$. When $C$ is misaligned (right), the difference becomes positive\ie the opposite behavior. Results are averaged over six random seeds.
  }
  \label{fig:acc_vs_p_and_weight_deltas}
\end{figure}
\subsection{Performance on Synthetic \& Real-World Datasets}
For comparing final performances, we report top-1 retrieval accuracy on the Synthetic-\texttt{XNOR} (\Cref{fig:acc_vs_p_and_weight_deltas}) and the three real-world trimodal datasets (\Cref{tab:results}).
Across all datasets, Gated Symile yields the best performance compared to Symile and CLIP.
The largest gain occurs on Synthetic-\texttt{XNOR} (from $0.3310$ to $0.8733$), consistent with the benchmark design in which exactly one non-target modality is intermittently misleading and the gate can suppress the unreliable factor before the multiplicative interaction.
For the \ac{ukb}, Gated Symile improves over Symile from $0.6570\pm0.012$ to $0.6819\pm0.010$, indicating that modulating modality contributions remains beneficial in a heterogeneous real-world cohort.
On Symile-MIMIC, the improvement is smaller but consistent (from $0.4556\pm0.006$ to $0.4670\pm0.005$), and gating does not degrade performance. 
On \ac{ukb}-Union, Gated Symile improves top-1 accuracy from $0.5278\pm0.009$ to $0.6000\pm0.007$. 
Despite the larger dataset, we do not observe a performance boost.
Instead, the additional non-target samples introduce greater variability, increasing the risk of overfitting. 
However, this setting highlights the benefit of adaptive gating when modalities are missing.
Overall, our results suggest that Gated Symile improves retrieval accuracy.
The strongest benefits appear in settings in which selective suppression is advantageous.

\begin{table}[t]
    \centering
    \caption{
        Analysis of mean gate statistics for the non-target modalities. Rows correspond to datasets and columns to gate diagnostics computed for the non-target modalities $m$ (\nth{2} column) relative to the target modality $t$. 
        We highlight $\cos(e_m^G, e_m)$: The relative changes across datasets mirror downstream performance\ie larger gains are associated with stronger deviations from the original embedding, reflected by smaller cosine similarity. This underscores the gate's effectiveness.
        %
        %
    }
    \label{tab:weight_cossims}
    \small
    \begin{tabular}{@{}l c c c c@{}}
    \toprule
    
    \textbf{Dataset} & 
    \textbf{Modalities} & 
    \bm{$w_{t\to m}$} &  
    \bm{$\cos(e^G_m, e_m)$} &
    \bm{$\cos(e^G_m, n_m)$} \\
    
    \midrule
    
    Synthetic-\texttt{XNOR} 
        & B \hspace{0.05cm} / \hspace{0.05cm} C
        & $0.3428$ \hspace{0.05cm} / \hspace{0.05cm} $0.1599$ 
        & $0.4385$ \hspace{0.05cm} / \hspace{0.05cm} $0.1656$ 
        & $0.7171$ \hspace{0.05cm} / \hspace{0.05cm} $0.9581$ \\

    Symile-MIMIC \cite{Saporta_Puli_Goldstein_Ranganath_2024} 
        & Lab. \hspace{0.05cm} / \hspace{0.05cm} ECG
        & $0.3656$ \hspace{0.05cm} / \hspace{0.05cm} $0.4568$ 
        & $0.9366$ \hspace{0.05cm} / \hspace{0.05cm} $0.9465$ 
        & $0.3596$ \hspace{0.05cm} / \hspace{0.05cm} $0.4837$ \\
        
    \ac{ukb} \cite{Sudlow_Gallacher_Allen_Beral_Burton_Danesh_Downey_Elliott_Green_Landray_etal_2015} 
        & Metabol. \hspace{0.05cm} / \hspace{0.05cm} \ac{ehr}
        & $0.5679$ \hspace{0.05cm} / \hspace{0.05cm} $0.3867$ 
        & $0.8921$ \hspace{0.05cm} / \hspace{0.05cm} $0.8203$ 
        & $0.6541$ \hspace{0.05cm} / \hspace{0.05cm} $0.7451$\\
        
    \ac{ukb}-Union \cite{Sudlow_Gallacher_Allen_Beral_Burton_Danesh_Downey_Elliott_Green_Landray_etal_2015} 
        & Metabol. \hspace{0.05cm} / \hspace{0.05cm} \ac{ehr}
        & $0.5808$ \hspace{0.05cm} / \hspace{0.05cm} $0.4884$ 
        & $0.7819$ \hspace{0.05cm} / \hspace{0.05cm} $0.6489$ 
        & $0.4854$ \hspace{0.05cm} / \hspace{0.05cm} $0.6408$\\
    
    \bottomrule
\end{tabular}
\end{table}
\begin{figure}[t]
  \centering
  
  \begin{subfigure}{0.48\linewidth}
    \centering
    \includegraphics[width=\linewidth]{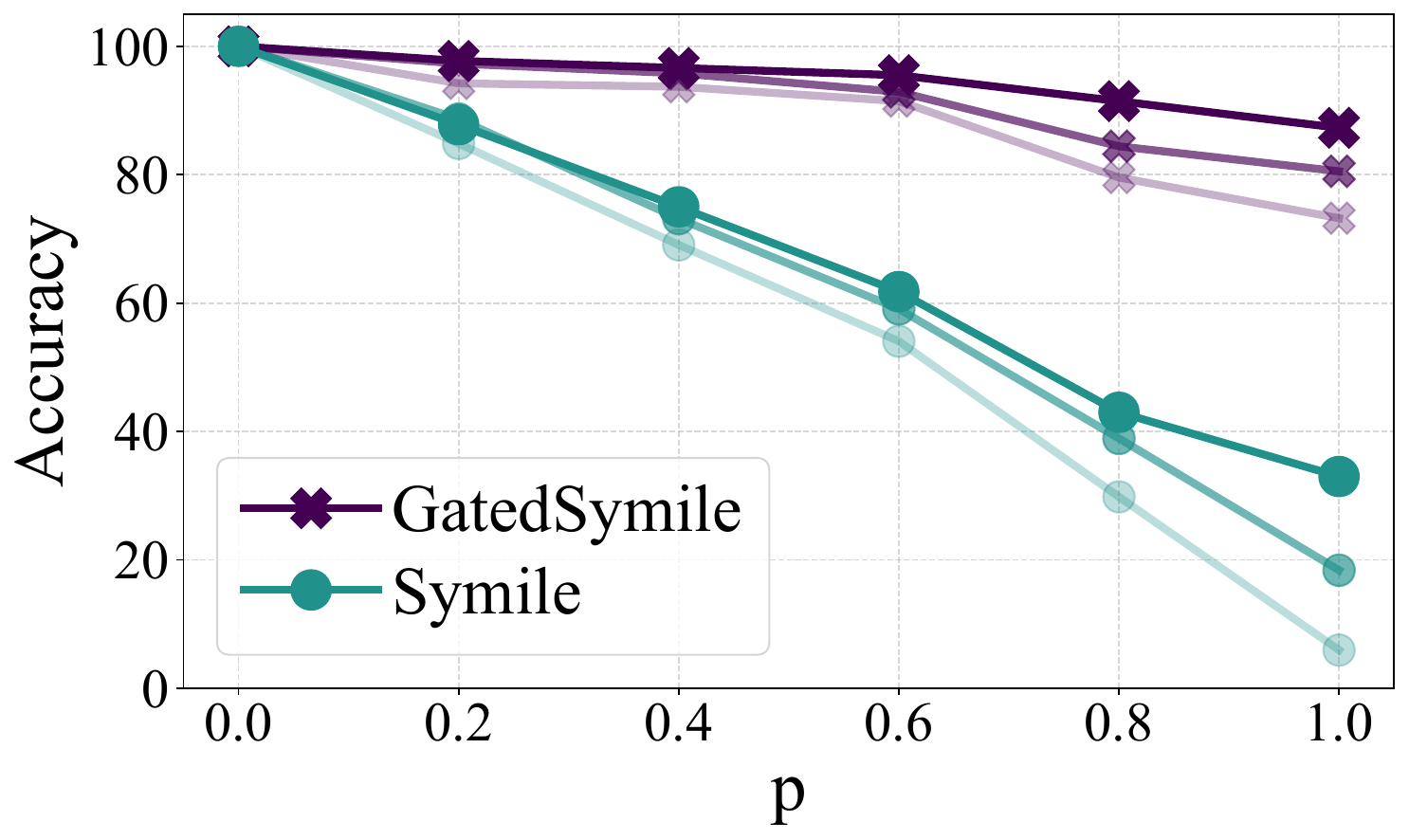}
  \end{subfigure}\hfill
  \begin{subfigure}{0.48\linewidth}
    \centering
    \includegraphics[width=\linewidth]{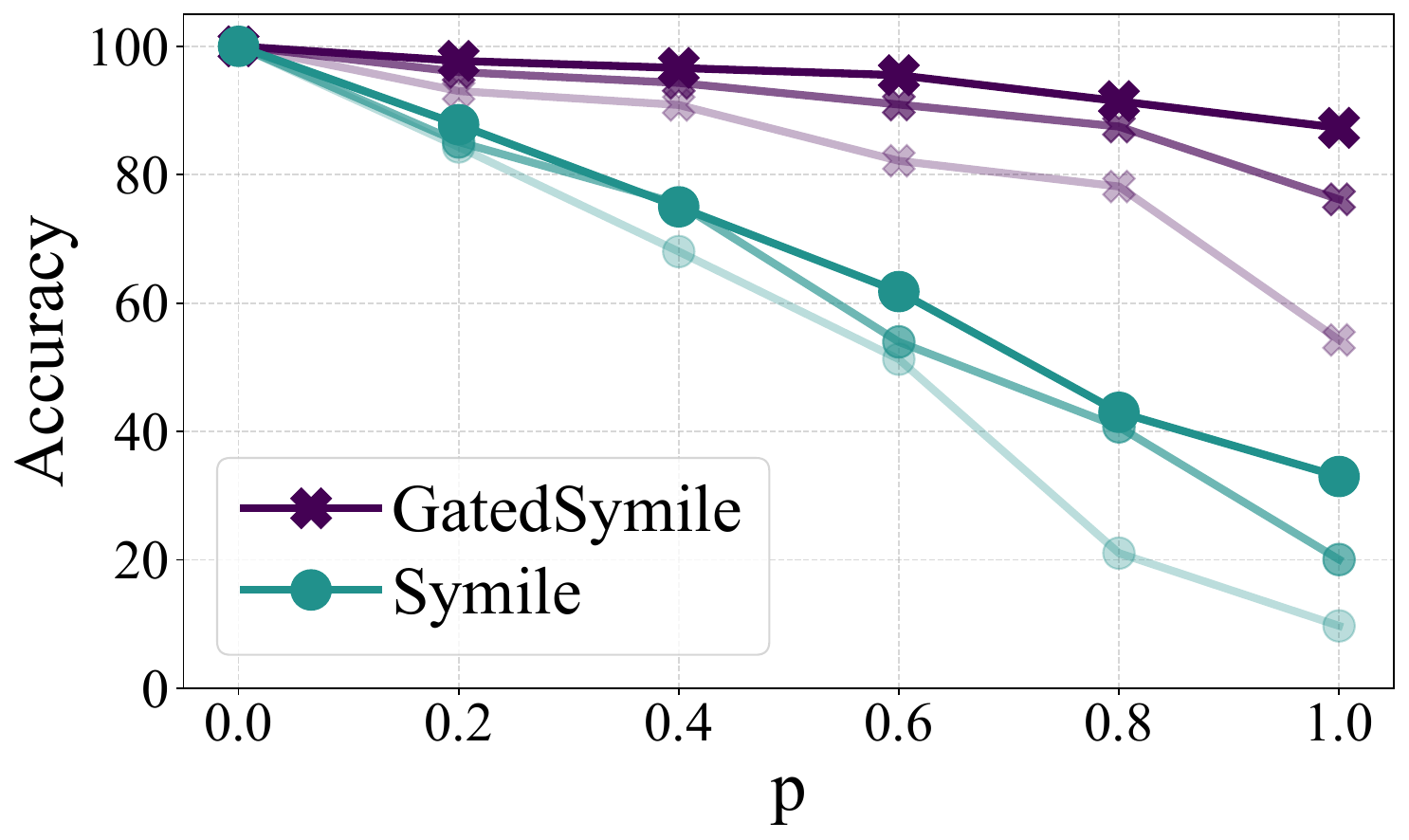}
  \end{subfigure}
  
  \caption{
    Scaling analyses of well-tuned models on the Synthetic-\texttt{XNOR} dataset under both increasing misalignment probability $p$ and batch sizes ($128$, $256$, $512$ illustrated with decreasing brightness).
    \textbf{(Left)} Joint scaling of $B$ and negatives per anchor $K$. Symile degrades substantially under misalignment, whereas Gated Symile remains markedly more robust, indicating increased fragility at larger contrastive scales.
    \textbf{(Right)} Scaling of $B$ while keeping negatives per anchor $K$ constant. This mirrors the joint-scaling regime, suggesting that the dominant scaling pathology is driven by the enlarged candidate pool: as $B$ grows, fragility becomes more prevalent.
    Gating mitigates this effect by suppressing unreliable factors, thereby preserving retrieval performance as scale increases.
    }
  \label{fig:acc_vs_p_scaling_b_and_bk}
\end{figure}
\subsection{Alignment, Weight and Embedding Analyses}
In the follwing, we analyze how gating reacts to unreliable modalities (\Cref{fig:acc_vs_p_and_weight_deltas,fig:acc_vs_p_scaling_b_and_bk,tab:weight_cossims}). 
In general, explainability approaches are well established \cite{Zhou_Khosla_Lapedriza_Oliva_Torralba_2015,Rheude_Wirtz_Kuijper_Wesarg_2024,Lundberg_Lee_2017,Sundararajan_Taly_Yan_2017,Vaswani_Shazeer_Parmar_Uszkoreit_Jones_Gomez_Kaiser_Polosukhin_2017}.
However, a growing body of work highlights the limitations of explainability approaches \cite{Jain_Wallace_2019,Jose_2025,Lipton_2018,Rudin_2019,Kindermans_Hooker_Adebayo_Alber_Schutt_Dahne_Erhan_Kim_2019,Adebayo_Gilmer_Muelly_Goodfellow_Hardt_Kim_2018,Sixt_Granz_Landgraf_2020}. 
Here, gate weights and embeddings are not considered interpretable quantities.
However, they can still provide coarse, aggregate trends that are useful for analysis.

\paragraph{Behavior Under Modality Misalignment}
We analyze misalignment conditions on the Synthetic-\texttt{XNOR} dataset (\Cref{fig:acc_vs_p_and_weight_deltas}).
As the misalignment probability $p$ increases, baseline \ac{sota} methods degrade, whereas Gated Symile remains near-ceiling accuracy.
This indicates that suppressing unreliable modalities prevents the \ac{mip} from collapsing under misleading inputs.
Further, we report the mean weight difference $w_{A\rightarrow B}-w_{A\rightarrow C}$ conditioned on which non-target modality is misaligned.
When $B$ is misaligned, the difference is negative (the gate assigns smaller weight to $B$ than to $C$), and when $C$ is misaligned the difference is positive.
This shows that the gate consistently shifts emphasis toward the reliable modality.
Moreover, there is a small trend that this signed difference increases with $p$, consistent with the gate making stronger decisions as misalignment increases.

\paragraph{Weight-Based vs. Representation-Based Diagnostics}
We compare two complementary gate diagnostics: mean gate weights $w_{t\to m}$ and cosine-based measures of representation change, $\cos(e_m^{G},e_m)$ and $\cos(e_m^{G},n_m)$ (\Cref{tab:weight_cossims}).
The mean weights alone can be hard to interpret because they average over heterogeneous, sample-dependent decisions and are influenced by remaining gate features.
Cosine similarities, in contrast, directly quantify whether the gate edits a modality embedding.
On Symile-MIMIC, where improvements are smallest, $\cos(e_m^{G},e_m)\approx 1$ and $\cos(e_m^{G},n_m)$ remains relatively low, indicating that the gate leaves embeddings largely unchanged.
On the \ac{ukb}, the gate shows moderate editing and a higher neutral-direction cosine which matches its intermediate performance gain.
For the \ac{ukb}-Union, $\cos(e_m^{G},e_m)$ decreases further compared to \ac{ukb}, indicating stronger embedding edits, but $\cos(e_m^{G},n_m)$ is not larger, suggesting a richer transformation than simple interpolation toward the neutral direction.
Finally, on Synthetic-\texttt{XNOR}, where gating yields the largest gains, $\cos(e_m^{G},e_m)$ is substantially reduced and $\cos(e_m^{G},n_m)$ is high for at least one non-target modality, consistent with pushing unreliable inputs toward the neutral direction.

\paragraph{Scaling Fragility and Gating Robustness}
Larger batch sizes typically improve contrastive learning performance \cite{Chen_Zhang_Xu_Chen_Duan_Chen_Tran_Zeng_Chilimbi_2022,Cheng_Zhang_Li_Leng_Hu_Wu_Zhao_Li_Bing_2024}.
However, in the presence of modality misalignment we observe the opposite trend (\Cref{fig:acc_vs_p_scaling_b_and_bk}): as the batch size $B$ and candidate pool grow, performance degrades more sharply with increasing misalignment probability $p$.
This appears both when jointly scaling $B$ and the number of negatives per anchor $K$, and when increasing $B$ alone while keeping $K$ fixed.
This suggests that the pathology is driven by the enlarged candidate pool.
In contrast, Gated Symile remains more stable across scaling regimes, indicating that suppressing unreliable modalities mitigates the fragility.

\subsection{Ablation Study}
\label{sec:ablation}
\begin{wraptable}[13]{r}{0.47\textwidth}
    \centering
    \vspace{-0.45cm}
    \caption{
        Well-tuned ablation of the gate on the \ac{ukb} (mean $\pm$ \ac{se}, details in \Cref{app:ablation_tuning}).
    }
    \label{tab:ablation}
    \small
    
    \begin{tabular}{@{}l c@{}}
    \toprule
    
    \textbf{Ablation} & 
    \textbf{Top-1 Accuracy $\uparrow$} \\
    
    \midrule

    Gated Symile 
        & $0.6819 \pm 0.010$ \\
        
    \textit{w/} neutral ones 
        & $0.6708 \pm 0.014$ \\
        
    \textit{w/o} \texttt{NULL} option
        & $0.6644 \pm 0.013 $ \\

    \textit{w/} neutral frozen
        & $0.6629 \pm 0.013$ \\
        
    \textit{w/} softmax (\textit{w/o} sigmoid)
        & $0.6622 \pm 0.012$ \\

    \textit{w/o} renorm 
        & $0.6578 \pm 0.013$ \\

    \textit{w/o} gate (Symile, pair)
        & $0.6570 \pm 0.012$ \\
        
    \textit{w/o} attention (\textit{w/} matrix)
        & $0.6446 \pm 0.014$ \\
        
    \textit{w/o} gate (Symile, n) 
        & $0.6419 \pm 0.024$ \\
        
    \textit{w/o} neutral \& renorm
        & $0.6314 \pm 0.014$ \\
    
    \bottomrule
    \end{tabular}
\end{wraptable}
We report a retuned and cross-validated ablation of the proposed gate on the \ac{ukb} (\Cref{tab:ablation}, Details in \Cref{app:ablation_tuning}).
Overall, the attention-based sigmoid gate with the \texttt{NULL} option and trainable neutral directions performs best, indicating that candidate-dependent suppression and an explicit neutral fallback are both important.  

\paragraph{Neutral Directions and Renormalization}
Using a fixed neutral direction (all-ones or frozen random), remains competitive but trails the model.
The all-ones variant is the next-best option, which is consistent with the intuition that the \ac{mip} contribution of a modality becomes weak in this case.
Dropping renormalization reduces accuracy, and removing both neutral interpolation and renormalization yields the worst gated variant.
This aligns with the view that unconstrained magnitudes can exacerbate multiplicative effects rather than suppress them.

\paragraph{Attention, Matrix, Sigmoid and Softmax Gating}
We find that sigmoid gating outperforms the softmax alternative.
Sigmoid allows multiple modalities to be weighted simultaneously, whereas softmax enforces competition. 
Finally, the matrix-based (candidate-independent) gate can be harmful, implying that static global weights are insufficient to capture sample-dependent misalignment.

\paragraph{Sampling Choice}
Interestingly, the \emph{pair} sampling slightly outperforms the \emph{n} sampling.
\emph{Pair} sampling draws $K$ negatives per anchor from the global candidate pool, which can be larger and more diverse than the in-batch shuffle used by \emph{n} sampling.
We therefore interpret the \emph{pair} vs \emph{n} comparison as both an efficiency and negative-diversity trade-off rather than a pure architectural ablation.

\section{Conclusion \& Future Work}
\label{sec:conclusion}
We studied robustness in multimodal contrastive learning beyond the bimodal setting and identified a failure mode of Symile-style objectives based on multiplicative interactions: misalignment in a single non-target modality can propagate through product terms and distort training. 
To address this, we proposed Gated Symile, a candidate-conditioned gating mechanism that adaptively downweights unreliable modalities.
The introduced gating mechanism interpolates embeddings toward learnable neutral directions and allows a \texttt{NULL} option when the target embedding indicates that reliable cross-modal alignment is unlikely.
Across a synthetic benchmark and three real-world trimodal retrieval datasets, Gated Symile improves top-1 retrieval over ungated Symile and well-tuned \ac{sota} baselines.
Our analyses revealed that the gate provides useful aggregate signals about modality reliability under misalignment.
Future work includes transferring the robustness induced by gating back into the encoders to study downstream tasks beyond retrieval and a deeper mechanistic interpretability analysis of the gate. 
\newpage 

\paragraph{Acknowledgement}
The authors acknowledge the Scientific Computing of the IT Division at the Charité Universitätsmedizin Berlin for providing computational resources that have contributed to the research results reported in this paper. This research has been conducted using the UK Biobank Resource under application number 49966.

\bibliography{neurips}
\bibliographystyle{plainnat}

\newpage 
\appendix

\section{Broader Impact and Ethics}
\label{sec:societal_impact}
We do not see clear societal impact concerns raised by the paper itself. The work is methodological and evaluates robustness in multimodal contrastive learning, with the stated goal of making learning under imperfect modalities more reliable rather than enabling a new high-risk application. 

\section{Relation to the Cauchy-Schwarz Bound}
\label{app:cauchy_schwarz_derivation}
To quantify the sensitivity of the \ac{mip} critic to corruption in a single modality, we compare its score on a clean tuple and on a corrupted tuple and study the score deviation $\Delta g := g_{\mathrm{corr}} - g_{\mathrm{clean}}$.
This difference isolates the effect of the corruption and admits a simple closed form because the \ac{mip} is multilinear in its arguments.
Starting from
\begin{equation}
    g_{\mathrm{corr}} - g_{\mathrm{clean}}
    =
    \frac{1}{\tau_{\text{MIP}}}\sum_{j=1}^{D}
    e_{t,j}
    \Bigg(\prod_{\substack{m=1\\ m\neq t,c}}^{M} e_{m,j}\Bigg)\delta_j,
    \label{eq:app_mip_delta_sum}
\end{equation}
define the vector
\begin{equation}
    a \;:=\; e_t \odot \prod_{\substack{m=1\\ m\neq t,c}}^{M} e_m \in \mathbb{R}^{D},
    \qquad
    \text{i.e.,}\quad
    a_j = e_{t,j}\prod_{\substack{m=1\\ m\neq t,c}}^{M} e_{m,j}.
    \label{eq:app_def_a}
\end{equation}
Then \Cref{eq:app_mip_delta_sum} can be written as an inner product,
\begin{align}
    g_{\mathrm{corr}} - g_{\mathrm{clean}}
    &=
    \frac{1}{\tau_{\text{MIP}}}\sum_{j=1}^{D} a_j\,\delta_j \\
    &=
    \frac{1}{\tau_{\text{MIP}}}\langle a,\delta\rangle.
    \label{eq:app_mip_delta_ip}
\end{align}
Taking absolute values yields
\begin{equation}
    \big|g_{\mathrm{corr}} - g_{\mathrm{clean}}\big|
    =
    \frac{1}{\tau_{\text{MIP}}}\,|\langle a,\delta\rangle|,
    \label{eq:app_abs}
\end{equation}
where we use $\tau_{\text{MIP}}>0$. By the Cauchy--Schwarz inequality, $|\langle a,\delta\rangle| \le \|a\|_2\,\|\delta\|_2$.
Since $\tau_{\text{MIP}}>0$, multiplying both sides by $1/\tau_{\text{MIP}}$ preserves the inequality direction, and thus
\begin{equation}
    \big|g_{\mathrm{corr}} - g_{\mathrm{clean}}\big|
    \le
    \frac{1}{\tau_{\text{MIP}}}\,\|a\|_2\,\|\delta\|_2
    =
    \frac{1}{\tau_{\text{MIP}}}\,
    \|\delta\|_2 \,
    \Big\|\, e_t \odot \!\!\!\prod_{\substack{m=1\\ m\neq t,c}}^{M}\!\!\! e_m \Big\|_2.
    \label{eq:app_cs_bound}
\end{equation}
Here $\prod e_m$ denotes elementwise multiplication across modalities, i.e.,
\begin{equation}
    \Big(\prod_{\substack{m=1\\ m\neq t,c}}^{M} e_m\Big)_j
    =
    \prod_{\substack{m=1\\ m\neq t,c}}^{M} e_{m,j}.
\end{equation}
Applying Cauchy-Schwarz yields a worst-case upper bound on corruption-induced score distortion, separating the perturbation magnitude $\|\delta\|_2$ from a multiplicative amplification term $\big\|e_t \odot \prod_{\substack{m=1\\ m\neq t,c}}^{M} e_m\big\|_2$.

\section{Gated MIP Sensitivity Derivation}
\label{app:gated_mip_sensitivity}

We derive the perturbation bound for the gated \ac{mip} analogously to \Cref{sec:method}.
For clarity, we consider the pre-normalized gated embedding. The normalization step is used in the implementation to keep embedding magnitudes comparable.

Recall from \Cref{eq:neutral_direction_interpolation,eq:gated_embedding} that
\begin{equation}
\tilde e_m = w_{t\to m} e_m + (1-w_{t\to m})n_m,
\end{equation}
and
\begin{equation}
e_m^G = (1-\alpha)e_m + \alpha \tilde e_m.
\end{equation}
Substituting $\tilde e_m$ gives
\begin{align}
e_m^G
&= (1-\alpha)e_m
+ \alpha\big(w_{t\to m}e_m+(1-w_{t\to m})n_m\big) \\
&= \big(1-\alpha+\alpha w_{t\to m}\big)e_m
+ \alpha(1-w_{t\to m})n_m.
\end{align}
We define the effective coefficient on the original embedding as
\begin{equation}
\beta_{t\to m}=1-\alpha+\alpha w_{t\to m}.
\end{equation}
Then the coefficient on the neutral direction is
\begin{align}
1-\beta_{t\to m}
&=1-\big(1-\alpha+\alpha w_{t\to m}\big) \\
&=\alpha-\alpha w_{t\to m} \\
&=\alpha(1-w_{t\to m}).
\end{align}
Hence, the gated embedding can be written equivalently as
\begin{equation}
e_m^G
=
\beta_{t\to m}e_m
+
(1-\beta_{t\to m})n_m.
\end{equation}

If the sigmoid-based gate with the \texttt{NULL} option is active, the effective non-target gate weight becomes
\begin{equation}
\bar w_{t\to m}=(1-p_{\mathrm{null}})w_{t\to m}.
\end{equation}
In this case, the interpolation is
\begin{equation}
\tilde e_m
=
\bar w_{t\to m}e_m
+
(1-\bar w_{t\to m})n_m,
\end{equation}
and therefore
\begin{align}
e_m^G
&=
(1-\alpha)e_m
+
\alpha\big(\bar w_{t\to m}e_m+(1-\bar w_{t\to m})n_m\big) \\
&=
\big(1-\alpha+\alpha \bar w_{t\to m}\big)e_m
+
\alpha(1-\bar w_{t\to m})n_m.
\end{align}
Thus, with the \texttt{NULL} option, we define
\begin{equation}
\beta_{t\to m}
=
1-\alpha+\alpha(1-p_{\mathrm{null}})w_{t\to m}.
\end{equation}
Again, the neutral coefficient is the complement:
\begin{align}
1-\beta_{t\to m}
&=
1-\big(1-\alpha+\alpha(1-p_{\mathrm{null}})w_{t\to m}\big) \\
&=
\alpha\big(1-(1-p_{\mathrm{null}})w_{t\to m}\big).
\end{align}
Therefore, both with and without the \texttt{NULL} option, the pre-normalized gated embedding can be written as
\begin{equation}
e_m^G
=
\beta_{t\to m}e_m
+
(1-\beta_{t\to m})n_m,
\end{equation}
where $\beta_{t\to m}$ denotes the effective coefficient on the original embedding.

Now suppose that a non-target modality $c\neq t$ is perturbed as $\hat e_c = e_c+\delta$ (\Cref{eq:mip_corruption_delta}), while the gate weights, $\beta_{t\to c}$, and the neutral direction $n_c$ are held fixed. Then
\begin{align}
\hat e_c^G
&=
\beta_{t\to c}\hat e_c
+
(1-\beta_{t\to c})n_c \\
&=
\beta_{t\to c}(e_c+\delta)
+
(1-\beta_{t\to c})n_c \\
&=
\beta_{t\to c}e_c
+
\beta_{t\to c}\delta
+
(1-\beta_{t\to c})n_c \\
&=
e_c^G
+
\beta_{t\to c}\delta.
\end{align}

The gated Symile score is
\begin{equation}
g^G(x^{(1)},\dots,x^{(M)})
=
\frac{1}{\tau_{\text{MIP}}}
\sum_{j=1}^{D}
\prod_{m=1}^{M} e^G_{m,j}.
\end{equation}
Therefore, the difference between the corrupted and clean gated scores is
\begin{align}
g^G_{\mathrm{corr}} - g^G_{\mathrm{clean}}
&=
\frac{1}{\tau_{\text{MIP}}}
\sum_{j=1}^{D}
e^G_{t,j}
\Bigg(
\prod_{\substack{m=1\\m\neq t,c}}^{M}
e^G_{m,j}
\Bigg)
\big(\hat e^G_{c,j}-e^G_{c,j}\big) \\
&=
\frac{\beta_{t\to c}}{\tau_{\text{MIP}}}
\sum_{j=1}^{D}
e^G_{t,j}
\Bigg(
\prod_{\substack{m=1\\m\neq t,c}}^{M}
e^G_{m,j}
\Bigg)
\delta_j.
\end{align}
Writing this as an inner product yields
\begin{equation}
g^G_{\mathrm{corr}} - g^G_{\mathrm{clean}}
=
\frac{\beta_{t\to c}}{\tau_{\text{MIP}}}
\left\langle
\delta,\,
e_t^G \odot
\!\!\!\prod_{\substack{m=1\\m\neq t,c}}^{M}\!\!\!
e_m^G
\right\rangle.
\end{equation}
Applying Cauchy--Schwarz gives
\begin{equation}
|g^G_{\mathrm{corr}} - g^G_{\mathrm{clean}}|
\le
\frac{\beta_{t\to c}}{\tau_{\text{MIP}}}
\|\delta\|_2
\Bigg\|
e_t^G \odot
\!\!\!\prod_{\substack{m=1\\m\neq t,c}}^{M}\!\!\!
e_m^G
\Bigg\|_2.
\end{equation}

For the basic gate, $\alpha\in[0,1]$ and $w_{t\to c}\in[0,1]$ imply
\begin{equation}
\beta_{t\to c}=1-\alpha+\alpha w_{t\to c}\in[1-\alpha,1].
\end{equation}
With the sigmoid \texttt{NULL} option, the effective weight is
$\bar w_{t\to c}=(1-p_{\mathrm{null}})w_{t\to c}$, hence
\begin{equation}
\beta_{t\to c}
=
1-\alpha+\alpha(1-p_{\mathrm{null}})w_{t\to c}
\in[1-\alpha,1].
\end{equation}
Increasing $p_{\mathrm{null}}$ reduces $\bar w_{t\to c}$ and therefore reduces $\beta_{t\to c}$. Thus, downweighting an unreliable modality contracts the effective perturbation entering the multiplicative score.
\section{Compute Environment}
\label{app:compute_environment}
\newacronym{hpc}{HPC}{High-Performance Cluster}
Our experiments are conducted on a \ac{hpc} with the following environment:
\begin{itemize}
    \item 21 Dell PowerEdge R7525 compute nodes, each with 64 AMD Epyc cores (Rome), 512GB RAM and 1 NVIDIA A100 40G GPU
    \item 2 Dell PowerEdge XE8545 compute nodes, each with 128 AMD Epyc cores (Milan), 512GB RAM, 4 NVIDIA A100 40G and 4 NVIDIA A100 80G GPUs (NVLink-connected)
\end{itemize}
\section{Additional Details}
\label{app:additionals}
In the following, we provide additional details for our proposed method, implementations, and comparisons.

\paragraph{MIP Normalization}
Following standard practice in contrastive learning, we use a learned logit scale to control the sharpness of the softmax over candidates \cite{Radford_Kim_Hallacy_Ramesh_Goh_Agarwal_Sastry_Askell_Mishkin_Clark_2021}.
Concretely, we parameterize the scale as $s=\exp(\gamma)>0$ and form logits $L=s\cdot S$ from a raw score matrix $S$.
For Symile-style objectives, $S$ is given by the \ac{mip} critic \cite{Saporta_Puli_Goldstein_Ranganath_2024}, whose variance increases with embedding dimension $d$ and number of modalities $M$ due to multiplicative interactions.
To stabilize early training and make temperature initialization comparable across $(d,M)$, we additionally apply a fixed $(d,M)$-dependent normalization to the raw \acp{mip} (a variance-style scaling analogous to variance-preserving initialization schemes \cite{Glorot_Bengio_2010,He_Zhang_Ren_Sun_2015}).
After this normalization, we multiply by the learned scale $s$ and apply cross-entropy.


\section{Hyperparameter Tuning}
\label{app:hyperparameter_tuning}
We maximize the validation retrieval accuracy by using Bayesian optimization without incorporating the batch size \citep{tuningplaybookgithub}. The methods are swept with $100$ runs. For experiments on the Synthetic-\texttt{XNOR} dataset, hyperparameters are retuned\eg for different values of $p$. For the UKB-Union results, sweep runs are reduced to $50$ due to longer runtimes. \Cref{lst:symile_mimic,lst:synthetic_xnor,lst:ukb,lst:clip_symile,lst:gated_symile} show the search spaces with respect to methods and datasets.

\begin{listing}[ht]
\begin{minted}[
    framesep=2mm,
    linenos,
    fontsize=\scriptsize,
]
{yaml}
method: bayes
metric:
  name: val/max_acc_top1
  goal: maximize

modelname.emb_dim:
    values: [1024]  # initially tuned from 256-8196
modelname.embedding_norm:
    values: [True]
# Encoders fixed to ResNets + MLP analogous to Saporta et al.
\end{minted}
\caption{Hyperparameters related to Symile-MIMIC.}
\label{lst:symile_mimic}
\end{listing}

\begin{listing}[ht]
\begin{minted}
[
    framesep=2mm,
    linenos,
    fontsize=\scriptsize,
]
{yaml}
method: bayes
metric:
  name: val/max_acc_top1
  goal: maximize

modelname.emb_dim:
    values: [256]  # initially tuned from 32-1024
modelname.embedding_norm:
    values: [True]
# Encoders fixed to MLPs
\end{minted}
\caption{Hyperparameters related to Synthetic-\texttt{XNOR}.}
\label{lst:synthetic_xnor}
\end{listing}

\begin{listing}[ht]
\begin{minted}
[
    framesep=2mm,
    linenos,
    fontsize=\scriptsize,
]
{yaml}
method: bayes
metric:
  name: val/max_acc_top1
  goal: maximize

modelname.emb_dim:
    values: [6144]  # initially tuned from 256-8196
modelname.embedding_norm:
    values: [True]
encoders.nmr.mlp.hidden_dims:
    values: [[1024,2048,4096]]  # initially tuned with 128-4096
encoders.nmr.mlp.hidden_dropouts:
    values: [[0.2,0.2,0.2]]  # initially tuned with 0.0-0.6
encoders.ehr.mlp.hidden_dims:
    values: [[1024,2048,4096]]  # initially tuned with 128-4096
encoders.ehr.mlp.hidden_dropouts:
    values: [[0.6,0.6,0.6]]  # initially tuned with 0.0-0.6
encoders.olink.mlp.hidden_dims:
    values: [[1024,2048,4096]]  # initially tuned with 128-4096
encoders.olink.mlp.hidden_dropouts:
    values: [[0.4,0.4,0.4]]  # initially tuned with 0.0-0.6
\end{minted}
\caption{Hyperparameters related to the \ac{ukb}.}
\label{lst:ukb}
\end{listing}

\begin{listing}[ht]
\begin{minted}
[
    framesep=2mm,
    linenos,
    fontsize=\scriptsize,
]
{yaml}
method: bayes
metric:
  name: val/max_acc_top1
  goal: maximize

modelname.logit_scale_init:
    min: -3
    max: 0
    distribution: "uniform"
optimizer.lr:
    min: 0.00001
    max: 0.01
    distribution: "log_uniform_values"
optimizer.warmup_steps:
    values: [0, 10, 50, 100, 200, 500, 1000, 1200]
optimizer.weight_decay:
    values: [0, 0.1, 0.01, 0.001]
\end{minted}
\caption{Hyperparameters related to Clip, Triangle, Gram and Symile.}
\label{lst:clip_symile}
\end{listing}

\begin{listing}[ht]
\begin{minted}
[
    framesep=2mm,
    linenos,
    fontsize=\scriptsize,
]
{yaml}
method: bayes
metric:
  name: val/max_acc_top1
  goal: maximize

modelname.logit_scale_init:
    min: -3
    max: 0
    distribution: "uniform"
modelname.gate_strength_init:
    min: -1
    max: 6
    distribution: "uniform"
modelname.neutral_type:
    values: ["random_trainable"]
modelname.gate_mode:
    values: ["attention"]
modelname.use_gate:
    values: [True]
modelname.use_null:
    values: [True]
modelname.renormalize:
    values: [True]
modelname.gate_type:
    values: ["sigmoid"]
modelname.gate_temp:
    min: 0.2
    max: 1.2
    distribution: "uniform"
optimizer.lr_gate_mul:
    min: 1.0
    max: 20.0
    distribution: "log_uniform_values"
modelname.gate_d_k:
    values: [1024, 3072, 6144]
optimizer.lr:
    min: 0.00001
    max: 0.01
    distribution: "log_uniform_values"
optimizer.warmup_steps:
    values: [0, 10, 50, 100, 200, 500, 1000, 1200]
optimizer.weight_decay:
    values: [0, 0.1, 0.01, 0.001]
\end{minted}
\caption{Hyperparameters related to Gated Symile.}
\label{lst:gated_symile}
\end{listing}

\section{Details About the Ablation}
\label{app:ablation_tuning}

\paragraph{Ablation Details}
We primarily conduct ablations of the proposed gating mechanism on \ac{ukb} without missing modalities. This choice is motivated by two factors: first, performance gains on Symile-MIMIC are marginal, making detailed ablations less informative. Second, the scale of \ac{ukb}-Union with missing modalities renders exhaustive ablation and retuning computationally prohibitive. While synthetic datasets do not fully reflect real-world conditions, we additionally report a re-tuned ablation on Synthetic-\texttt{XNOR} in \Cref{tab:ablation_xnor} to provide a controlled analysis of the method.  
\begin{table}[h] 
    \centering
    \vspace{-0.45cm}
    \caption{
        Well-tuned ablation of the gate on the Synthetic-\texttt{XNOR} dataset ordered ablation-wise according to the ablation on the \ac{ukb}.
    }
    \label{tab:ablation_xnor}
    \small
    
    \begin{tabular}{@{}l c@{}}
    \toprule
    
    \textbf{Ablation} & 
    \textbf{Top-1 Accuracy $\uparrow$} \\
    
    \midrule

    Gated Symile
        & $0.8730$ \\

    \textit{w/} neutral ones
        & $0.8523$ \\
        
    \textit{w/o} \texttt{NULL} option
        & $0.8920$ \\

    \textit{w/} neutral frozen
        & $0.8690$ \\
        
    \textit{w/} softmax (\textit{w/o} sigmoid)
        & $0.8813$ \\

    \textit{w/o} renorm
        & $0.8736$ \\

    \textit{w/o} gate (Symile, pair)
        & $0.3251$ \\
        
    \textit{w/o} attention (\textit{w/} matrix)
        & $0.2380$ \\
        
    \textit{w/o} gate (Symile, n)
        & $0.3043$ \\
        
    \textit{w/o} neutral \& renorm
        & $0.4917$ \\
    
    \bottomrule
    \end{tabular}
\end{table}

\paragraph{Ablation Hyperparameter Tuning}
Ablation studies can be misleading if components are removed while keeping the original hyperparameters fixed: changing the model\eg removing a gate, \texttt{NULL}, renormalization, or attention, can substantially shift the optimal learning rate, regularization, temperature, and even effective capacity, so performance differences may reflect suboptimal tuning rather than the true contribution of the ablated component \cite{Rheude_Eils_Wild_2025,Dodge_Gururangan_Card_Schwartz_Smith_2019}. To avoid conflating architectural changes with mismatched hyperparameters, we re-run dataset-specific hyperparameter tuning for every ablation and report the best-performing configuration under the same validation protocol and search budget (\Cref{app:ablation_hyperparameters_full,app:ablation_hyperparameters_ones,app:ablation_hyperparameters_nonull,app:ablation_hyperparameters_neutral_frozen,app:ablation_hyperparameters_softmax,tab:ablation_hyperparameters_renorm,app:ablation_hyperparameters_matrix,app:ablation_hyperparameters_noneutral}). The ablation experiments are swept with $50$ runs. Parameter counts are listed in \Cref{tab:ablation_params}.

\begin{table}[h]
    \centering
    \caption{
        Parameter counts of (Gated) Symile and our ablated variants. Gate-related parameters (without encoder parameters) per ablation configuration are listed, so they may change across variants.
    }
    \small
    \begin{tabular}{@{}l c@{}}
        \toprule
    
        \textbf{Ablation} &  
        \textbf{Parameters $\downarrow$} \\
        
        \midrule
    
        Gated Symile 
            & $132$M \\
            
        \textit{w/} neutral ones 
            & $44.1$M \\
            
        \textit{w/o} \texttt{NULL} option
            & $264$M \\
    
        \textit{w/} neutral frozen
            & $44.1$M \\
            
        \textit{w/} softmax (\textit{w/o} sigmoid)
            & $44.1$M \\
    
        \textit{w/o} renorm 
            & $264$M \\
    
        \textit{w/o} gate (Symile, pair)
            & $0.0$ \\
            
        \textit{w/o} attention (\textit{w/} matrix)
            & $18.4$K \\
            
        \textit{w/o} gate (Symile, n) 
            & $0.0$ \\
            
        \textit{w/o} neutral \& renorm
            & $264$M \\
        
        \bottomrule
    \end{tabular}
    \label{tab:ablation_params}
\end{table}

\begin{table}[h]
    \centering
    \caption{Ablation hyperparameters: Gated Symile.}
    \label{app:ablation_hyperparameters_full}
    \small
    \begin{tabular}{@{}l r@{}}
    \toprule
    \textbf{Parameter} & \textbf{Value} \\
    \midrule

    \texttt{modelname.negative\_sampling} & \texttt{pair} \\

    \texttt{modelname.emb\_dim} & \texttt{6144} \\
    \texttt{modelname.logit\_scale\_init} & \texttt{-0.0273882549} \\

    \texttt{optimizer.lr} & \texttt{0.0009146280} \\
    \texttt{optimizer.warmup\_steps} & \texttt{1200} \\
    \texttt{optimizer.weight\_decay} & \texttt{0.01} \\
    \texttt{optimizer.lr\_gate\_mul} & \texttt{18.0142950406} \\

    \texttt{modelname.use\_gate} & \texttt{True} \\
    \texttt{modelname.gate\_d\_k} & \texttt{3072} \\
    \texttt{modelname.gate\_mode} & \texttt{attention} \\
    \texttt{modelname.gate\_strength\_init} & \texttt{5.1367568069} \\
    \texttt{modelname.gate\_temp} & \texttt{0.2859855525} \\
    \texttt{modelname.gate\_type} & \texttt{sigmoid} \\

    \texttt{modelname.neutral\_type} & \texttt{random\_trainable} \\

    \bottomrule
    \end{tabular}
\end{table}

\begin{table}[h]
    \centering
    \caption{Ablation hyperparameters: \textit{w/} neutral ones.}
    \label{app:ablation_hyperparameters_ones}
    \small
    \begin{tabular}{@{}l r@{}}
    \toprule
    \textbf{Parameter} & \textbf{Value} \\
    \midrule

    \texttt{modelname.negative\_sampling} & \texttt{pair} \\

    \texttt{modelname.emb\_dim} & \texttt{6144} \\
    \texttt{modelname.logit\_scale\_init} & \texttt{-0.4172494091} \\

    \texttt{optimizer.lr} & \texttt{0.0008030235} \\
    \texttt{optimizer.warmup\_steps} & \texttt{1200} \\
    \texttt{optimizer.weight\_decay} & \texttt{0.0} \\
    \texttt{optimizer.lr\_gate\_mul} & \texttt{11.8582226203} \\

    \texttt{modelname.use\_gate} & \texttt{True} \\
    \texttt{modelname.gate\_d\_k} & \texttt{1024} \\
    \texttt{modelname.gate\_mode} & \texttt{attention} \\
    \texttt{modelname.gate\_strength\_init} & \texttt{5.9809122372} \\
    \texttt{modelname.gate\_temp} & \texttt{0.8945044902} \\
    \texttt{modelname.gate\_type} & \texttt{sigmoid} \\

    \texttt{modelname.neutral\_type} & \texttt{ones} \\
    \texttt{modelname.renormalize} & \texttt{True} \\
    \texttt{modelname.use\_null} & \texttt{True} \\

    \bottomrule
    \end{tabular}
\end{table}

\begin{table}[h]
    \centering
    \caption{Ablation hyperparameters: \textit{w/o} \texttt{NULL} 
    \label{app:ablation_hyperparameters_nonull}option.}
    \small
    \begin{tabular}{@{}l r@{}}
    \toprule
    \textbf{Parameter} & \textbf{Value} \\
    \midrule

    \texttt{modelname.negative\_sampling} & \texttt{pair} \\

    \texttt{modelname.emb\_dim} & \texttt{6144} \\
    \texttt{modelname.logit\_scale\_init} & \texttt{-0.0385663517} \\

    \texttt{optimizer.lr} & \texttt{0.0024638659} \\
    \texttt{optimizer.warmup\_steps} & \texttt{500} \\
    \texttt{optimizer.weight\_decay} & \texttt{0.001} \\
    \texttt{optimizer.lr\_gate\_mul} & \texttt{5.3507888856} \\

    \texttt{modelname.use\_gate} & \texttt{True} \\
    \texttt{modelname.gate\_d\_k} & \texttt{6144} \\
    \texttt{modelname.gate\_mode} & \texttt{attention} \\
    \texttt{modelname.gate\_strength\_init} & \texttt{5.0979182757} \\
    \texttt{modelname.gate\_temp} & \texttt{0.4696580431} \\
    \texttt{modelname.gate\_type} & \texttt{sigmoid} \\

    \texttt{modelname.neutral\_type} & \texttt{random\_trainable} \\
    \texttt{modelname.renormalize} & \texttt{True} \\
    \texttt{modelname.use\_null} & \texttt{False} \\

    \bottomrule
    \end{tabular}
\end{table}

\begin{table}[h]
    \centering
    \caption{Ablation hyperparameters: \textit{w/} neutral frozen.}
    \label{app:ablation_hyperparameters_neutral_frozen}
    \small
    \begin{tabular}{@{}l r@{}}
    \toprule
    \textbf{Parameter} & \textbf{Value} \\
    \midrule

    \texttt{modelname.negative\_sampling} & \texttt{pair} \\

    \texttt{modelname.emb\_dim} & \texttt{6144} \\
    \texttt{modelname.logit\_scale\_init} & \texttt{-0.0885577025} \\

    \texttt{optimizer.lr} & \texttt{0.0007128068} \\
    \texttt{optimizer.warmup\_steps} & \texttt{1200} \\
    \texttt{optimizer.weight\_decay} & \texttt{0.001} \\
    \texttt{optimizer.lr\_gate\_mul} & \texttt{11.8457842440} \\

    \texttt{modelname.use\_gate} & \texttt{True} \\
    \texttt{modelname.gate\_d\_k} & \texttt{1024} \\
    \texttt{modelname.gate\_mode} & \texttt{attention} \\
    \texttt{modelname.gate\_strength\_init} & \texttt{5.5512603864} \\
    \texttt{modelname.gate\_temp} & \texttt{0.7620343329} \\
    \texttt{modelname.gate\_type} & \texttt{sigmoid} \\

    \texttt{modelname.neutral\_type} & \texttt{random\_frozen} \\
    \texttt{modelname.renormalize} & \texttt{True} \\
    \texttt{modelname.use\_null} & \texttt{True} \\

    \bottomrule
    \end{tabular}
\end{table}

\begin{table}[h]
    \centering
    \caption{Ablation hyperparameters: \textit{w/} softmax.}
    \label{app:ablation_hyperparameters_softmax}
    \small
    \begin{tabular}{@{}l r@{}}
    \toprule
    \textbf{Parameter} & \textbf{Value} \\
    \midrule

    \texttt{modelname.negative\_sampling} & \texttt{pair} \\

    \texttt{modelname.emb\_dim} & \texttt{6144} \\
    \texttt{modelname.logit\_scale\_init} & \texttt{-0.0981230686} \\

    \texttt{optimizer.lr} & \texttt{0.0027758740} \\
    \texttt{optimizer.warmup\_steps} & \texttt{1000} \\
    \texttt{optimizer.weight\_decay} & \texttt{0.001} \\
    \texttt{optimizer.lr\_gate\_mul} & \texttt{1.0920241972} \\

    \texttt{modelname.use\_gate} & \texttt{True} \\
    \texttt{modelname.gate\_d\_k} & \texttt{1024} \\
    \texttt{modelname.gate\_mode} & \texttt{attention} \\
    \texttt{modelname.gate\_strength\_init} & \texttt{5.3640459076} \\
    \texttt{modelname.gate\_temp} & \texttt{0.5666661356} \\
    \texttt{modelname.gate\_type} & \texttt{softmax} \\

    \texttt{modelname.neutral\_type} & \texttt{random\_trainable} \\
    \texttt{modelname.renormalize} & \texttt{True} \\
    \texttt{modelname.use\_null} & \texttt{True} \\

    \bottomrule
    \end{tabular}
\end{table}

\begin{table}[h]
    \centering
    \caption{Ablation hyperparameters: \textit{w/o} renorm.}
    \label{tab:ablation_hyperparameters_renorm}
    \small
    \begin{tabular}{@{}l r@{}}
    \toprule
    \textbf{Parameter} & \textbf{Value} \\
    \midrule
    
    \texttt{modelname.negative\_sampling} & \texttt{pair} \\
    
    \texttt{modelname.emb\_dim} & \texttt{6144} \\
    \texttt{modelname.logit\_scale\_init} & \texttt{-0.0676587788} \\
    \texttt{optimizer.lr} & \texttt{0.0033153970} \\
    \texttt{optimizer.warmup\_steps} & \texttt{1000} \\
    \texttt{optimizer.weight\_decay} & \texttt{0.01} \\
    \texttt{modelname.use\_gate} & \texttt{True} \\
    \texttt{modelname.gate\_d\_k} & \texttt{6144} \\
    \texttt{modelname.gate\_mode} & \texttt{attention} \\
    \texttt{modelname.gate\_strength\_init} & \texttt{5.1129839132} \\
    \texttt{modelname.gate\_temp} & \texttt{1.0882940326} \\
    \texttt{modelname.gate\_type} & \texttt{sigmoid} \\
    \texttt{modelname.neutral\_type} & \texttt{random\_trainable} \\
    \texttt{modelname.renormalize} & \texttt{False} \\
    \texttt{modelname.use\_null} & \texttt{True} \\
    \texttt{optimizer.lr\_gate\_mul} & \texttt{1.1935684118} \\
    \bottomrule
    \end{tabular}
\end{table}

\begin{table}[h]
    \centering
    \caption{Ablation hyperparameters: \textit{w/o} attention.}
    \label{app:ablation_hyperparameters_matrix}
    \small
    \begin{tabular}{@{}l r@{}}
    \toprule
    \textbf{Parameter} & \textbf{Value} \\
    \midrule

    \texttt{modelname.negative\_sampling} & \texttt{pair} \\

    \texttt{modelname.emb\_dim} & \texttt{6144} \\
    \texttt{modelname.logit\_scale\_init} & \texttt{-0.1154340629} \\

    \texttt{optimizer.lr} & \texttt{0.0026584110} \\
    \texttt{optimizer.warmup\_steps} & \texttt{1200} \\
    \texttt{optimizer.weight\_decay} & \texttt{0.01} \\
    \texttt{optimizer.lr\_gate\_mul} & \texttt{2.4905551621} \\

    \texttt{modelname.use\_gate} & \texttt{True} \\
    \texttt{modelname.gate\_d\_k} & \texttt{3072} \\
    \texttt{modelname.gate\_mode} & \texttt{matrix} \\
    \texttt{modelname.gate\_strength\_init} & \texttt{1.2150563177} \\
    \texttt{modelname.gate\_temp} & \texttt{0.5117133726} \\
    \texttt{modelname.gate\_type} & \texttt{sigmoid} \\

    \texttt{modelname.neutral\_type} & \texttt{random\_trainable} \\

    \bottomrule
    \end{tabular}
\end{table}

\begin{table}[h]
    \centering
    \caption{Ablation hyperparameters: \textit{w/o} neutral \& random.}
    \label{app:ablation_hyperparameters_noneutral}
    \small
    \begin{tabular}{@{}l r@{}}
    \toprule
    \textbf{Parameter} & \textbf{Value} \\
    \midrule

    \texttt{modelname.negative\_sampling} & \texttt{pair} \\

    \texttt{modelname.emb\_dim} & \texttt{6144} \\
    \texttt{modelname.logit\_scale\_init} & \texttt{-0.1298500657} \\

    \texttt{optimizer.lr} & \texttt{0.0012920771} \\
    \texttt{optimizer.warmup\_steps} & \texttt{1200} \\
    \texttt{optimizer.weight\_decay} & \texttt{0.01} \\
    \texttt{optimizer.lr\_gate\_mul} & \texttt{9.8383592465} \\

    \texttt{modelname.use\_gate} & \texttt{True} \\
    \texttt{modelname.gate\_d\_k} & \texttt{6144} \\
    \texttt{modelname.gate\_mode} & \texttt{attention} \\
    \texttt{modelname.gate\_strength\_init} & \texttt{0.0530650431} \\
    \texttt{modelname.gate\_temp} & \texttt{0.2552342123} \\
    \texttt{modelname.gate\_type} & \texttt{sigmoid} \\

    \texttt{modelname.neutral\_type} & \texttt{None} \\
    \texttt{modelname.renormalize} & \texttt{True} \\
    \texttt{modelname.use\_null} & \texttt{True} \\

    \bottomrule
    \end{tabular}
\end{table}


\end{document}